\documentclass[10pt,twocolumn,letterpaper]{article}
\usepackage{cvpr}              

\usepackage[dvipsnames, table]{xcolor}
\usepackage{tabularx}
\usepackage{algorithm}
\usepackage{mathabx}
\usepackage{amsmath}
\DeclareMathOperator{\sign}{sign}
\usepackage{algpseudocode}
\usepackage{pifont}
\usepackage{float}
\usepackage{graphicx}
\usepackage{subcaption}
\usepackage{adjustbox}
\usepackage{afterpage}
\usepackage{lipsum} 
\usepackage{booktabs} 
\usepackage{siunitx}  
\newcommand{\cmark}{\ding{51}}
\newcommand{\xmark}{\ding{55}}

\definecolor{cvprblue}{rgb}{0.21,0.49,0.74}
\usepackage[pagebackref,breaklinks,colorlinks,citecolor=cvprblue]{hyperref}
\def\paperID{2869} 
\def\confName{CVPR}
\def\confYear{2024}
\title{ExMap: Leveraging Explainability Heatmaps for Unsupervised Group Robustness to Spurious Correlations}

\author{Rwiddhi Chakraborty,
 Adrian Sletten,
 Michael C. Kampffmeyer
\\
Department of Physics and Technology, UiT The Arctic University of Norway\\
{\tt \small firstname[.middle initial].lastname@uit.no}
}
 
\begin{document}
\maketitle
\begin{abstract}
Group robustness strategies aim to mitigate learned biases in deep learning models that arise from spurious correlations present in their training datasets. However, most existing methods rely on the access to the label distribution of the groups, which is time-consuming and expensive to obtain. As a result, unsupervised group robustness strategies are sought. Based on the insight that a trained model's classification strategies can be inferred accurately based on explainability heatmaps, we introduce ExMap, an unsupervised two stage mechanism designed to enhance group robustness in traditional classifiers. ExMap utilizes a clustering module to infer pseudo-labels based on a model's explainability heatmaps, which are then used during training in lieu of actual labels. Our empirical studies validate the efficacy of ExMap - We demonstrate that it bridges the performance gap with its supervised counterparts and outperforms existing partially supervised and unsupervised methods. Additionally, ExMap can be seamlessly integrated with existing group robustness learning strategies. Finally, we demonstrate its potential in tackling the emerging issue of multiple shortcut mitigation\footnote{Code available at \url{https://github.com/rwchakra/exmap}}.
\end{abstract}    
\section{Introduction}
\label{sec:intro}

\begin{figure*}[!t]
    \centering
    \includegraphics[width=\linewidth]{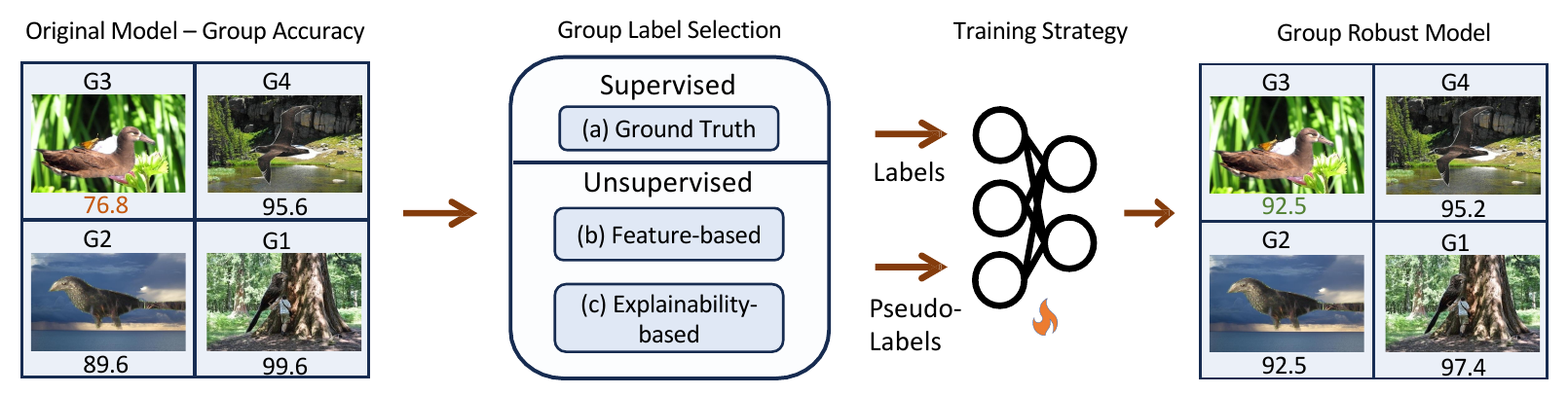}\vspace{-2mm}
    \caption{To improve the original models worst group accuracy, most current approaches rely on supervised group labels (a), which requires extensive annotation processes. Unsupervised approaches have relied on extracting pseudo labels based on the models feature representations (b), where information can be highly entangled. ExMap instead infers group pseudo labels based on explainability heatmaps (c), leading to improved worst group performance.}\vspace{-3mm}
    \label{fig:exmap_highlevel}
\end{figure*}

Deep neural network classifiers trained for classification tasks, have invited increased scrutiny from the research community due to their overreliance on spurious correlations present in the training data \cite{DBLP:journals/corr/abs-2110-04301, brendel2018approximating, Geirhos2018ImageNettrainedCA, Zhang2022, Bahng2019LearningDR}. This is related to the broader aspect of Shortcut Learning \cite{Geirhos2020ShortcutLI}, or the Clever Hans effect \cite{Lapuschkin2019UnmaskingCH}, where a model picks the path of least resistance to predict data, thus relying on shortcut features that are not causally linked to the label. The consequence of this phenomenon is that, although such models may demonstrate impressive mean accuracy on the test data, they may still fail on challenging subsets of the data, i.e. the groups \cite{duchi2023distributionally, duchi2018statistics, Oren2019DistributionallyRL}. As a result, group robustness is a natural objective to be met to mitigate reliance on spurious correlations. Thus, instead of evaluating models based on mean test accuracy, evaluating them on \textit{worst group accuracy} has been the recent paradigm~\cite{Izmailov2022OnFL, Nam2022SpreadSA, Zhou2021ExaminingAC, Lokhande2022TowardsGR}, resulting in the emergence of group robustness techniques. By dividing a dataset into pre-determined groups of spurious correlations, classifiers are then trained to maximize the \textit{worst group accuracy} - As a result, the spurious attribute that the model is most susceptible to is considered the shortcut of interest.

In Figure \ref{fig:exmap_highlevel}, we illustrate the group robustness paradigm. Given a dataset, a robustness strategy takes as input the group labels and retrains a base classifier (such as Expected Risk Minimization, i.e. ERM) to improve the worst group accuracy (G3 in this case). GroupDRO \cite{Sagawa2019DistributionallyRN}  was one of the early influential works that introduced the group robustness paradigm. Further, it demonstrated a strategy that could indeed improve worst group accuracy. One limitation of this approach was the reliance on group labels in the training data, which was replaced with the reliance on group labels in the validation data in successive works \cite{Liu2021JustTT, Kirichenko2022LastLR}. However, while these efforts have made strides in enhancing the accuracy of trained classifiers for underperforming groups, many hinge on the assumption that the underlying groups are known apriori and that the group labels are available, which is often impractical in real-world contexts. An unsupervised approach, as illustrated in Figure \ref{fig:exmap_highlevel}, would ideally estimate pseudo-labels that could be inputs to any robustness strategy, leading to improved worst group robustness. An example of such a fully unsupervised worst group robustness approach is (GEORGE) \cite{Sohoni2020NoSL}. GEORGE clusters the penultimate layer features in a UMAP reduced space, demonstrating impressive results on multiple datasets. In this work, we instead show that clustering \textit{explainability heatmaps} instead, is more beneficial in improving worst group robustness. Intuitively, this stems from the fact that a pixel-attribution based explainability method in input space focuses only on the relevant image features (pixel space) in the task, discarding other entangled features irrelevant for the final prediction.  

In our work, we circumvent the need for a group labeled dataset by introducing ExMap, a novel two stage mechanism: First, we extract explainability heatmaps from a trained (base)
model on the dataset of interest (we use the validation set \textit{without} group labels). Next, we use a clustering module to produce pseudo-labels for the validation data. The resulting pseudo-labels can then be used for any off-the-shelf group robustness learning strategy in use today. ExMap is also flexible in the choice of clustering algorithm. We show that attaching the ExMap mechanism to baseline methods leads to improved performance over the unsupervised counterparts, and  further closes the gap to supervised and partially supervised counterparts. Additionally, we demonstrate that ExMap is also useful in the recent \textit{multiple shortcut} paradigm \cite{Li2022AWD}, where current popular supervised approaches have been shown to struggle. We conclude with an extended analysis on why clustering explainability heatmaps is more beneficial than raw features. In summary, our contributions include:
\begin{itemize}
    \item [1.] ExMap: A simple but efficient unsupervised, strategy agnostic mechanism for group robustness that leverages explainability heatmaps and clustering to generate pseudo-labels for underlying groups. 
    \item [2.] An extended analysis that provides intuition and insight into why clustering explainability heatmaps leads to superior results over other group-robustness baseline methods.
    \item [3.]  Demonstrating the usefulness of ExMap in improving worst group robustness in both the single shortcut and multiple shortcut settings. 
\end{itemize}

\section{Related Work}
\label{sec:formatting}

\paragraph{Single shortcut mitigation with group labels}
The paradigm of taking a frozen base model and proposing a shortcut mitigation strategy to maximise \textit{worst group accuracy} was introduced in Group-DRO (gDRO) \cite{Sagawa2019DistributionallyRN}. However, the requirement of group labels in both training and validation data motivated the proposal of mitigation strategies without training labels. 
This has resulted partially supervised approaches~\cite{Sohoni2021BARACKPS} that only require a small set of group labels as well as in several methods that only require the validation group labels~\cite{Liu2021JustTT,Kirichenko2022LastLR,Nam2020LearningFF}. One such example is DFR\cite{Kirichenko2022LastLR}, which re-trains the final layer of a base ERM model on a balanced, reweighting dataset. Most relevant to our work, GEORGE~\cite{Sohoni2020NoSL} proposes an unsupervised mechanism to generate pseudo-labels for retraining by clustering raw features, and can therefore be considered the closest method to our proposed ExMap.
We show that clustering heatmaps is a more beneficial and intuitive technique for generating pseudo-labels, as attributing the model performance on the input data pixels leads to a more intuitive interpretation of which features are relevant for the task, and which are not. Our method, ExMap, leverages this insight and clusters the heatmaps instead, leading to improved performance over GEORGE and its two variants - GEORGE(gDRO) trained with the Group-DRO strategy, and GEORGE(DFR), trained with the DFR strategy. 
\vspace{-0.25cm}
\paragraph{Other Strategies for Shortcut Mitigation}
There are other extant works that mitigate spurious correlations without adopting the group-label based paradigm directly. MaskTune \cite{Taghanaki2022MaskTuneMS}, for instance, learns a mask over discriminatory features to reduce reliance on spurious correlations. CVar DRO \cite{Lvy2020LargeScaleMF} proposes an efficient robustness strategy using conditional value at risk (CVar). DivDis \cite{Lee2023DiversifyAD}, on the other hand, proposes to train multiple functions on source and target data, identifying the most informative subset of labels in the target data. Discover-and-Cure (DISC) \cite{Wu2023DiscoverAC} discovers spurious concepts using a predefined concept bank, then intervenes on the training data to mitigate the spurious concepts, while ULA \cite{Tsirigotis2023GroupRC} uses a pretrained self-supervised model to train a classifier to detect and mitigate spurious correlations. While these approaches do not directly adopt the group-label, we show that the proposed explainability heatmap-based approach is more efficient in improving the worst-group accuracy.

\paragraph{Multi-Shortcut Mitigation}
The single shortcut setting is a simpler benchmark as the label is spuriously correlated with only a single attribute. However, real world datasets are challenging, and may contain multiple spurious attributes correlated with an object of interest. As a result, when one spurious attribute is known, mitigating the reliance on this attribute may exacerbate the reliance on another. The recently introduced Whac-A-Mole \cite{Li2022AWD} dilemma for multiple shortcuts demonstrates this phenomenon with datasets containing multiple shortcuts (e.g. background and co-occurring object). Single shortcut methods fail to mitigate \textit{both} shortcuts at once, leading to a spurious conservation principle, where if one shortcut is mitigated, the other is exacerbated. The authors introduce Last Layer Ensemble (LLE) to mitigate multiple shortcuts in their datasets, by training a separate classifier for each shortcut. However, LLE's reliance on apriori knowledge of dataset shortcuts is impractical in the real world. We evaluate ExMap in this context and show that it is effective as an unsupervised group robustness approach to the multi-shortcut setting. 
\vspace{-0.25cm}
\paragraph{Heatmap-based Explainability}

The challenge of attributing learned features to the decision making of a model in the image space has a rich history. The techniques explored can be differentiated on a variety of axes. LIME, SHAP,  LRP \cite{Guo2017LIMELI, Lundberg2017AUA, Bach2015OnPE} are early model-agnostic methods, while Grad-CAM and Integrated Gradients\cite{Selvaraju2016GradCAMVE, Sundararajan2017AxiomaticAF} are gradient based attribution methods. We use LRP in this work owing to its popularity, but in principle, the heatmap extraction module can incorporate any other method widely in use today. LRP is a backward propagation based technique relying on the relevance conservation principle across each neuron in each layer. The output is a set of relevance scores that can be attributed to a pixel wise decomposition of the input image. Heatmap-based explainability techniques have also been used in conjunction with clustering, in the context of discovering model strategies for classification, and disparate areas such as differential privacy~\cite{Lapuschkin2019UnmaskingCH, Chen2018DetectingBA, Schulth2022DetectingBP}.

\section{Worst Group Robustness}
In this section, we provide notation and brief background of the group robustness problem. We are given a dataset \(D\) with image-label pairs being defined as \(D = \{(x_i, y_i)\}_{i=1}^{N}\), where \(x_i\) represents an image, \(y_i\) is its corresponding label, and \(N\) is the number of pairs in the dataset. The model's prediction for an image \(x\) is \(y_{\text{pred}} = \hat{f}(x)\). The cross-entropy loss for true label \(y\) and predicted label \(y_{\text{pred}}\) is given by \(L(y, y_{\text{pred}}) = -\sum_{c=1}^{C} y_c \log(y_{\text{pred},c})\), where \(C\) is the number of classes. Then, an ERM 
classifier simply minimizes the average loss over the training data:
\begin{equation}
\hat{f} = \arg\min_{f} \frac{1}{N} \sum_{i=1}^{N} L(y_i, f(x_i))
\end{equation}
where \(\hat{f}\) is the model obtained after training. Next, given the validation data $D$, we assume that for the class label set $L = \{c_1, c_2, ..., c_k\}$ there exists a corresponding spurious attribute set $A = \{a_1, a_2, ..., a_m\}$, such that the group label set $G: L \bigtimes A$. For example, in CelebA, typically $a:$ Gender (Male/Female), and $c$: Blonde Hair (Blonde/Not Blonde). In this case $L = \{0,1\}$, and $A = \{0, 1\}$. Then, the optimization can be described as the worst-expected loss over the validation set, conditioned on the group labels and the spurious attributes:

\begin{equation}
\hat{f}^* = \arg\min_{f} \max_{(c_i, a_j) \in G} \mathbb{E}_{(x, y) \in D} [L(y, f(x)) | c_i, a_j]
\end{equation}

As discussed before, recent works aim to design strategies over the (base) trained model to minimize this objective. For example, JTT collects an error set from the training data, and then upweights misclassified examples during the second training phase. DFR reweights the features responsible for misclassifications during the first phase in its finetuning stage. Note, however, that both these methods rely on the validation set \textit{group labels} $G_{val}$ to finetune the network. We consider the case where $G_{train} = \phi$ and $G_{val} = \phi$. We do not have access to group labels, and must therefore infer pseudo-labels in an unsupervised manner so that existing group robustness methods can be used.

\begin{figure*}[t]
    \centering    
    \includegraphics[width=\linewidth]{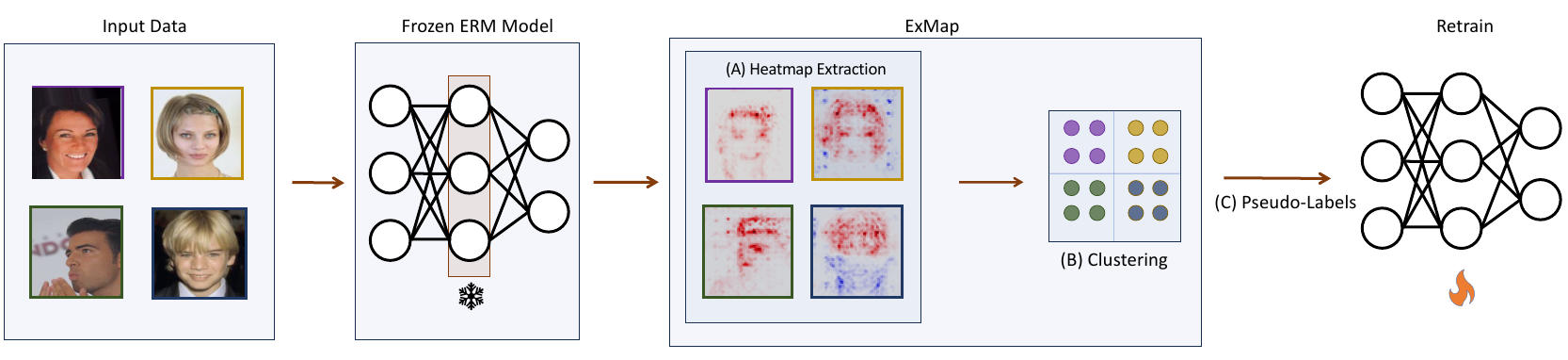}
    \caption{Our Proposed Method: ExMap  facilitates group-robustness by extracting explainability heatmaps from the frozen base ERM model for the validation data (A). These heatmaps are then clustered (B) to obtain pseudo-labels for the underlying groups, which are subsequently chosen for the retraining strategy (C).}\vspace{-3mm}
    \label{fig:exmap}
\end{figure*}
\section{Leveraging Explainability Heatmaps for Group Robustness -- ExMap}

In this section, we describe ExMap, an intuitive and efficient approach for unsupervised group robustness to spurious correlations. ExMap is a two-stage method, illustrated in Figure~\ref{fig:exmap}. In the first stage, we extract explainability heatmaps for the model predictions. In the second stage, we cluster the heatmaps to generate pseudo-labels. These pseudo-labels can then be used on any off-the-shelf group robustness strategy in use today. In our work, we demonstrate the strategy agnostic nature of ExMap by running it on two popular strategies - JTT and DFR.

\subsection{Explainability Heatmaps}
We use LRP \cite{Bach2015OnPE} in this stage to generate pixel attributions in the input space. This allows us to focus only on the relevant features for the task. Specifically, given the validation data \(D_{val} = \{(x_i, y_i)\}_{i=1}^{M}\), we use pixel wise relevance score  $r_x = (\mathcal{LRP}(x)) \forall x \in D_{val}$. Specifically, for each data point $x$, the relevance score is defined on a per-neuron-per layer basis. For an input neuron $n_k$ at layer $k$, and an output neuron $n_l$ at the following layer $l$, the relevance score $R_{k}$ is intuitively a measure of how much this particular input $n_k$ contributed to the output value $n_l$: 
\begin{equation}
     R_{k} = \sum_{l: k \xrightarrow{}l}\frac{z_{kl}}{z_l + \varepsilon \cdot \sign(z_{l})} 
     \label{eq:lrp}
\end{equation} 
where $z_{l} = n_kw_{kl}$ for the weight connection $w_{kl}$. $R_k$ is computed for all the neurons at layer $k$, and backpropagated from the output layer to the input layer to generate pixel level relevance scores $r_x$ for each data input $x$. We can thus build the heatmap set $R = \{r_x | x \in D_{val}\}$. This process is summarized in Algorithm ~\ref{algo:exmap}. 

\begin{figure*}[t]
    \centering
  \includegraphics[width=1.0\textwidth, trim={2cm 0 13cm 0},clip]{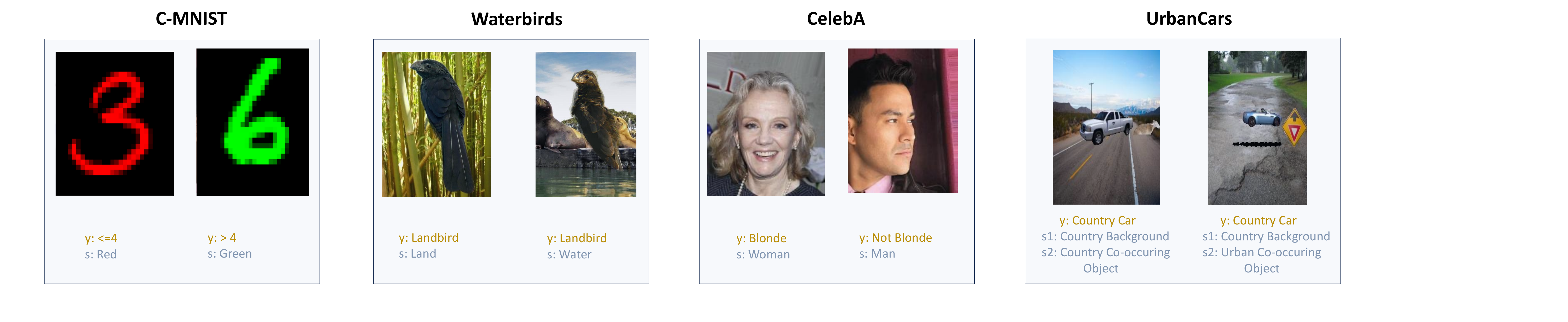}\vspace{-2mm}
    \caption{The datasets used in our work, visualized with respect to the class labels, and the shortcuts $s$. For the complete list of datasets and more details, please refer to the supplementary material.}\vspace{-4mm}
    \label{fig:datasets}
\end{figure*}
\label{sec:method}

\begin{algorithm}[t]
\caption{Generating Pseudo-labels using G-ExMap}
\label{algo:exmap}
\begin{algorithmic}[1]
\State \textbf{Input:} Dataset \(D_{val}\), ERM Model \(\mathcal{M}\), DataLoader \(\mathcal{L}\)
\State \textbf{Output:} Pseudo-labels \(\hat{G}\)

\Procedure{GeneratePseudoLabels}{$D, \mathcal{M}, \mathcal{L}$}

    \State \(R \gets \emptyset\)  
    \Comment{Initialize heatmap set}
    
    \For{each batch \(x\) in \(\mathcal{L}\)}
        \State \(\text{pred} \gets \arg\max_{i} \mathcal{M}(x)_i\)
        
        \For{each layer \(k, l\)}
            \State Compute \(z_{l} \gets n_kw_{kl}\)
        \EndFor
        \State Compute LRP relevance \(r_x\) for \(x\) using Eq. \ref{eq:lrp}
        
        \State Add \(r_x\) to \(R\)
    \EndFor
    \State Cluster \(R\) using G-ExMap method:
    \State \(\hat{A} \gets \text{Cluster}(R) \)
    \Comment{Estimated spurious labels}

    \State Combine class labels $L$ with $\hat{A}$
    \[ \hat{G} \gets L \times \hat{A} \]
    
    \State \Return \( \text{Pseudo-labels } \hat{G} \)

\EndProcedure
\end{algorithmic}
\end{algorithm}

\subsection{Clustering}

In the second stage, we cluster the LRP representations from the first stage. The intuition here is that over the data, the heatmaps capture the different strategies undertaken by the model for the classification task \cite{Lapuschkin2019UnmaskingCH}. The clustering module helps identify dominant model strategies used for the classification task. By identifying such strategies and resampling in a balanced manner, ExMap guides the model to be less reliant on the dominant features across the data, i.e. the spurious features. The heatmaps serve as an effective proxy to describe model focus areas. We have two options in choosing how to cluster: Local-ExMap (L-ExMap), where we cluster heatmaps on a per-class basis, and Global-ExMap (G-ExMap), where we cluster all the heatmaps at once, and segment by class labels. We present the G-ExMap results in this paper, owing to better empirical results. 

Specifically, given the Heatmap set $R$ as described in Algorithm \ref{algo:exmap}, the estimated spurious labels are generated by the global clustering method, $\hat{A} = \text{Cluster}(R)$
where \(\text{Cluster (.)}\) represents a clustering function.
Now, given class label set $L$ and estimated spurious label set $\hat{A}$, we can generate our pseudo-\textit{group} label set $\hat{G} = L \bigtimes A$ by selecting each $a_i \in \hat{A}$, and each $c_k \in L$, to create $\{c_k, a_i\}$ $\forall k, i$.

 In principle, it doesn't matter what clustering method we use, but that the clustering process itself outputs useful pseudo-labels. For our work, we leverage spectral clustering with an eigengap heuristic, inspired by SPRAY \cite{Lapuschkin2019UnmaskingCH}. Later, we show that the choice of the clustering method does not have a significant effect on the results. The outputs, which are the pseudo group labels for the validation data $D_{val}$, can now be used as labels in lieu of ground truth labels to train any group robustness strategy in use. Note how in principle, \textit{any} method that uses group labels (training or validation) would benefit from this approach. To apply this to the training set $D$, one would simply repeat Algorithm \ref{algo:exmap} on $D$. In this work, we apply ExMap to two common group robustness strategies - JTT \cite{Liu2021JustTT} and DFR \cite{Kirichenko2022LastLR}. Thus, we demonstrate the strategy-agnostic nature of our approach that can be applied to any off-the-shelf method using group labels today.

\section{Experiments}
\label{sec:experiments}

In this section we first present the datasets, baselines, and experimental setup. Next, we present the results and discussion\footnote{A discussion of the limitations and societal impact can be found in the supplementary material.}.

\paragraph{Datasets}

We use CelebA \cite{liu2015faceattributes}, Waterbirds \cite{Sagawa2019DistributionallyRN, Zhou2016PlacesAI, WahCUB_200_2011}, C-MNIST \cite{Arjovsky2019InvariantRM}, and Urbancars \cite{Li2022AWD}. In CelebA, the class label to be predicted is hair colour (Blonde/Not Blonde), and the spurious attribute is gender (Male/Female). For Waterbirds, the class label is the bird type (waterbird/landbird), and the spurious attribute is the background (land/water). In C-MNIST, the class label is if the number is smaller than or equal to four. Any number lesser than or equal to four is assigned blue, while all numbers greater than four are assigned the color red, with a correlation of 99\%. Thus, the spurious attribute is the color. 
For Urbancars, the class label is the car type (country/urban), and the spurious attributes are the background and co-occuring object (both country/urban). We create two variants of Urbancars: The first variant is Urbancars (BG), where only the background object is the spurious attribute. The second variant is Urbancars (CoObj), where the co-occuring object is the spurious attribute. We present single shortcut results on CelebA, Waterbirds, C-MNIST, Urbancars (BG) and Urbancars (CoObj). For the multiple shortcut setting, we use the original UrbanCars dataset with both shortcuts \cite{Li2022AWD}. An overview over the considered datasets can be found in Figure \ref{fig:datasets}. We present more dataset details in the supplementary. 

\paragraph{Baselines}

We use the unsupervised approaches DivDis, MaskTune, and two variants of GEORGE (with gDRO and DFR) as the baselines in our work. We also adapt LfF, JTT, and CVar DRO to the unsupervised setting as additional baselines. We train the ERM model using an Imagenet-pretrained Resnet-50, and use the open source implementations of the baselines to generate our results. Specifically, we implement GEORGE(DFR), ExMap, and JTT. Remaining results are reported from \cite{Taghanaki2022MaskTuneMS}, \cite{Liu2021JustTT}, and \cite{Lee2023DiversifyAD}. 

\paragraph{Setup}

We make sure to use the same hyperparameters from the baseline papers to reproduce the results. We utilise a composite of LRP rules to get the explainability heatmaps as recommended by \cite{Montavon2019LayerWiseRP, Kohlbrenner2019TowardsBP}. Following their recommendations we use LRP-$\epsilon$ for the dense layers near the output of the model with small epsilon ($\epsilon \ll 1$), followed by LRP-$\gamma$ for the convolutional layers.

For the spectral clustering, we use the affinity matrix, and cluster-QR \cite{Schulth2022DetectingBP} to perform the clustering. The eigengap heuristic is applied to the 10 smallest eigenvalues of the Laplacian matrix to select the number of significant clusters to use. We demonstrate later that using a simpler clustering approach such as k-means can also emit reasonable results. For more details on the affinity matrix, clustering and pseudo-label generation, please see the supplementary. 
\begin{table*}[!t]
    \centering
    \setlength{\tabcolsep}{12pt}
    \resizebox{0.9\linewidth}{!}{\begin{tabular}{lcllll}
    
        \toprule
        \textbf{Methods} & \textbf{Group Info} & \multicolumn{2}{c}{\textbf{Waterbirds}} & \multicolumn{2}{c}{\textbf{CelebA}} \\
        \cmidrule(lr){3-4} \cmidrule(lr){5-6}
        & Train/Val & WGA(\%)\color{Green}{$\uparrow$} & Mean(\%) & WGA(\%)\color{Green}{$\uparrow$} & Mean(\%) \\
        \midrule
        Base (ERM) & \xmark/\xmark & 76.8 & 98.1 & 41.1 & 95.9\\
        \hline
        \rowcolor{gray!25}
        Group DRO & \cmark/\cmark & 91.4 & 93.5 & 88.9 & 92.9 \\
        \rowcolor{gray!25}
        EIIL & \cmark/\cmark & 87.3 & 93.1 & 81.3 & 89.5 \\
        \rowcolor{gray!25}
        BARACK & \cmark/\cmark & 89.6 & 94.3 & 83.8 & 92.8 \\
        \rowcolor{gray!25}
        CVar DRO  & \xmark/\cmark & 75.9 & 96.0 &  64.4 & 82.5 \\
        \rowcolor{gray!25}
        LfF  & \xmark/\cmark & 78.0 & 91.2 &  77.2 & 85.1\\
        \rowcolor{gray!25}
        JTT  & \xmark/\cmark & 86.7 & 93.3 & 81.1 & 88.0 \\
        \rowcolor{gray!25}
        DFR & \xmark/\cmark & 92.1 & 96.7 & 86.9 & 91.1 \\

        \hline
        GEORGE (gDRO) & \xmark/\xmark & 76.2 & 95.7 & 53.7 & 94.6 \\
        CVar DRO  & \xmark/\xmark & 62.0 & 96.0 &  36.1 & 82.5 \\
        LfF  & \xmark/\xmark & 44.1 & 91.2 &  24.4 & 85.1 \\
        JTT  & \xmark/\xmark & 62.5 & 93.3 & 40.6 & 88.0 \\
        DivDis* & \xmark/\xmark & 81.0 & - & 55.0 & - \\ 
        MaskTune & \xmark/\xmark & 86.4 & 93.0 & 78.0 & 91.3 \\
        GEORGE (DFR) & \xmark/\xmark & 91.7 & 96.5 & 83.3 & 89.2 \\ 
        DFR+ExMap (ours) & \xmark/\xmark & \textbf{92.5} & 96.0 & \textbf{84.4} & 91.8 \\
        \bottomrule
    \end{tabular}}
    \caption{Worst group and mean accuracy on the test sets of the different datasets. The Group Info column showcases for each method whether group labels are used for that split of the data (\xmark = does not use group labels, \cmark = uses group labels). We report the average results over 5 runs after hyperparameter tuning. Gray rows represent supervised approaches. *DivDis does not report mean test accuracy results.}
    \label{tab:mainresult}
\end{table*}

\begin{table*}[!t]
    \centering
    \resizebox{0.9\linewidth}{!}{%
    \begin{tabular}{lcllllll}
        \toprule
        \textbf{Methods} & \textbf{Group Info} & \multicolumn{2}{c}{\textbf{C-MNIST}} & \multicolumn{2}{c}{\textbf{Urbancars (BG)}} & \multicolumn{2}{c}{\textbf{Urbancars (CoObj)}} \\
        \cmidrule(lr){3-4} \cmidrule(lr){5-6} \cmidrule(lr){7-8}
        & Train/Val & WGA(\%)\color{Green}{$\uparrow$} & Mean(\%) & WGA(\%)\color{Green}{$\uparrow$} & Mean(\%) & WGA(\%)\color{Green}{$\uparrow$} & Mean(\%) \\
        \midrule
        Base (ERM) & \xmark/\xmark & 39.6 & 99.3 & 55.6 & 90.2 & 50.8 & 92.7\\
        \rowcolor{gray!25}
        DFR & \xmark/\cmark & 74.2 & 93.7 & 77.5 & 81.0 & 84.7 & 88.2 \\ 
        \hline
        GEORGE (DFR) & \xmark/\xmark & 71.7 & 95.2 & 69.1 & 83.6 & 76.9 & 91.4 \\
        DFR+ExMap (ours) & \xmark/\xmark & \textbf{72.5} & 94.9 & \textbf{71.4} & 93.2 & \textbf{79.2} & 93.2 \\
        \bottomrule
    \end{tabular}}
    \caption{Worst Group accuracy and mean accuracy on C-MNIST, Urbancars (BG), and Urbancars (CoObj). We use GEORGE as the baseline, since both GEORGE and ExMap significantly outperform other unsupervised methods on Waterbirds and CelebA. Gray rows represent supervised approaches.}
    \label{tab:mainresult-george}
\end{table*}

\subsection{Results: Single Shortcut}

In Table \ref{tab:mainresult} we present the single shortcut results for the datasets. First, we note that with no supervision, ExMap based DFR improves significantly upon ERM. Second, we note the improved performance of ExMap based DFR over the unsupervised baselines, including GEORGE, our closest baseline.
Further, since the DFR-based GEORGE and ExMap significantly outperforms the other baselines, we present results comparing these two methods on C-MNIST, Urbancars (BG) and Urbancars (CoObj) in Table \ref{tab:mainresult-george}. In both tables, we demonstrate the superiority of clustering heatmaps to generate pseudo-labels instead of the raw features as in GEORGE. These results also show that the groups inferred by ExMap are indeed useful for worst group robustness to spurious correlations. Third, we note the gap in performance between DFR and ExMap based DFR. Since the former uses validation labels, we expect an increased accuracy, but we can report better performance on Waterbirds, and within 3\%\, 2\%\, 8\%\, and 6\%\ of the DFR results on the remaining datasets. On CelebA, our results are within 5\%\ of Group-DRO, which demonstrates the best overall results. However, note that Group-DRO is a fully supervised approach, using labels from both the training and validation sets. For all datasets, we are able to outperform GEORGE, our closest baseline. As discussed before, while mean accuracy is not the appropriate metric to track in the group robustness setting (ERM has the best overall mean accuracy but the worst overall worst group accuracy), we can still confirm that ExMap based DFR does not witness significant drops in performance. 

\subsection{Results: Multiple Shortcuts}

Here, we present the results on the UrbanCars data, which contains multiple shortcuts in the images - the background and the co-occurring object in the image. This dataset was introduced in the recent work on multiple shortcut mitigation \cite{Li2022AWD}, where the authors show that mitigating one shortcut may lead to a reliance on another shortcut in the data, rendering the single shortcut setting incomplete (the Whac-a-Mole problem). The authors introduce a new set of metrics for the task - The \textbf{BG Gap}, which is the drop in accuracy between mean and cases when only the background is uncommon, the \textbf{CoObj Gap} which is the drop in accuracy between mean and cases when only the co-occurring object is uncommon, and the \textbf{BG+CoObj Gap}, the drop when both the background and the co-occurring object are uncommon. A mitigation strategy should witness a smaller drop from the original accuracy when compared to others. In Table \ref{tab:multishortcut}, we present the ExMap based DFR results with respect to DFR, ERM, and GEORGE(DFR). We also present results of three variants of DFR: DFR (Both), which is retraining on the original UrbanCars data with both shortcuts. DFR(BG) retrains on UrbanCars with only the background shortcut, and DFR(CoObj) retrains with only the co-occuring object shortcut. Red values indicate an increase in gap when compared to ERM, which is undesirable (the Whac-A-Mole dilemma). Note that the first three DFR methods have access to the group labels, while GEORGE and ExMap do not. Table \ref{tab:multishortcut} demonstrates some important results: First, that DFR + ExMap consistently posts lower drops than the base ERM model. Second, that ExMap does not witness an increase in gap on any of the metrics compared to ERM, unlike GEORGE(DFR), which exhibits the Whac-A-Mole dilemma for the CoObj Gap. Finally, the DFR variants exhibit the Whac-A-Mole dilemma: For a DFR variant retrained on a particular shortcut, the reliance on that shortcut is mitigated (e.g. DFR (BG) mitigates the BG Gap), but the other shortcut reliance is exacerbated (DFR (BG) exhibits a higher CoObj Gap than ERM).
Note that DFR uses the validation group labels, and hence will be more useful in mitigating shortcuts than our unsupervised setting. In fact, as demonstrated in \cite{Li2022AWD}, training separate classifiers for each shortcut is the best approach to mitigating multiple shortcuts, which explains DFR's best overall results. However, this setting assumes availability to the shortcut labels, which ExMap does not assume. Yet, it demonstrates a robust performance for the multi-shortcut setting even in the unsupervised setting, outperforming GEORGE, its closest unsupervised competitor. 

\begin{table}[tbp]
\centering
\resizebox{.5\textwidth}{!}{\begin{tabular}{lllc}
\toprule
\textbf{Method} & \textbf{BG Gap} \color{Green}{$\uparrow$} & \textbf{CoObj Gap} \color{Green}{$\uparrow$} & \textbf{BG+CoObj Gap} \color{Green}{$\uparrow$}\\
\midrule
ERM & -8.2 & -14.2 & -69.0 \\
\rowcolor{gray!25}
DFR (Both) & -4.6 & -5.4 & -14.2 \\
\rowcolor{gray!25}
DFR (BG) & -0.3 & -29.2 (\textcolor{red}{$\times$ 2.06}) & -33.2 \\
\rowcolor{gray!25}
DFR (CoObj) & -16.3 (\textcolor{red}{$\times$ 1.99}) & -0.5 & -19.1 \\
\midrule
GEORGE (DFR) & -7.0 & -15.4 (\textcolor{red}{$\times$1.08}) & -63.4 \\
DFR+ExMap (ours) & \textbf{-5.9} & \textbf{-9.9} & \textbf{-30.7} \\
\bottomrule
\end{tabular}}
\vspace{-3mm}
\caption{Multiple Shortcuts on UrbanCars. Red values indicate the Whac-A-Mole dilemma: Mitigating one shortcut exacerbates reliance on the other (compared to ERM). ExMap proves to be the most robust in this setting, and outperforms GEORGE, its direct unsupervised counterpart.}\vspace{-4mm}
\label{tab:multishortcut}
\end{table}
\begin{figure}[!t]
\centering
\includegraphics[width=\linewidth]{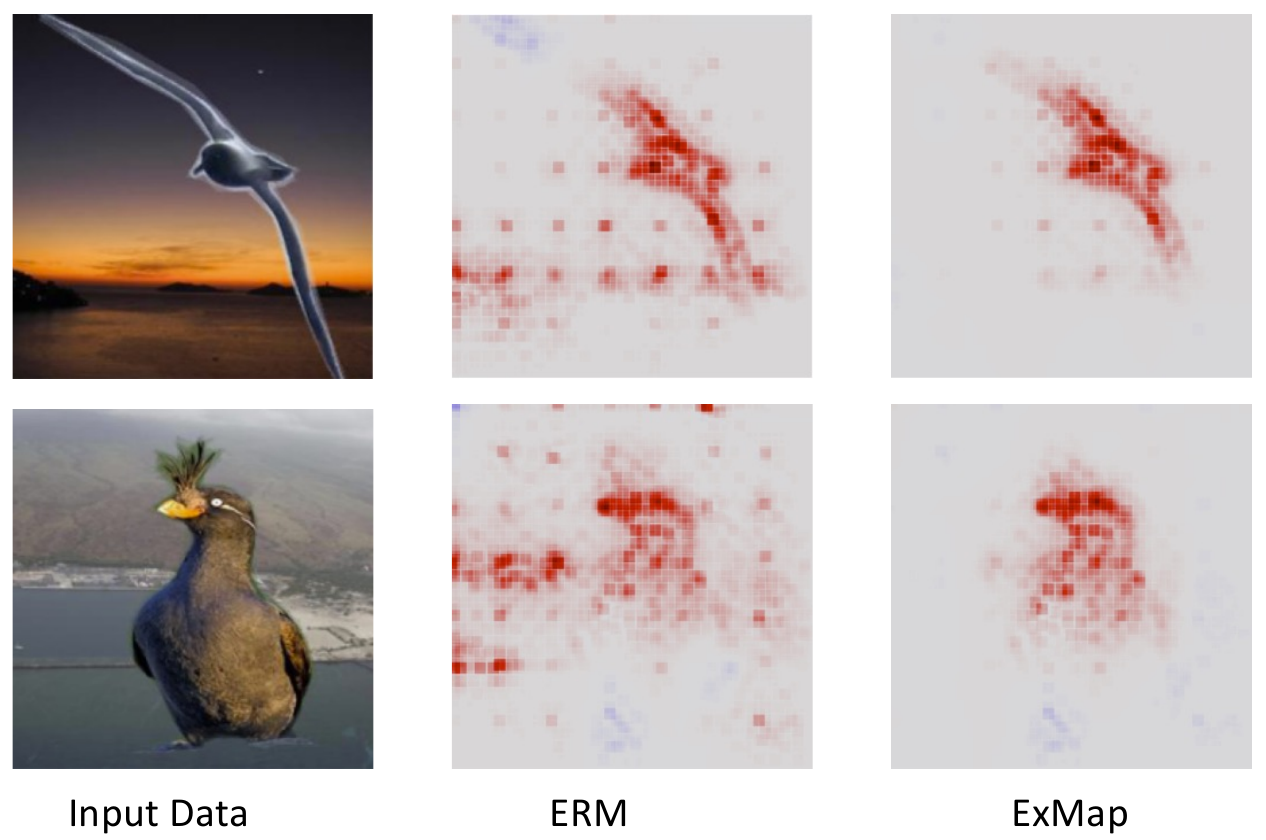}
    \caption{ERM and ExMap Heatmaps - Left: The Input images. Middle: ERM model explanations. Right: Improved group robustness using ExMap. Our method helps improve the focus on relevant attributes, in turn improving the pseudo-label estimation for retraining.}\vspace{-3mm}
    \label{fig:heatmaps}
\end{figure}

\section{Analysis}
\label{sec:analysis}
\begin{figure}[!t]
    \centering
    \includegraphics[width=\linewidth]{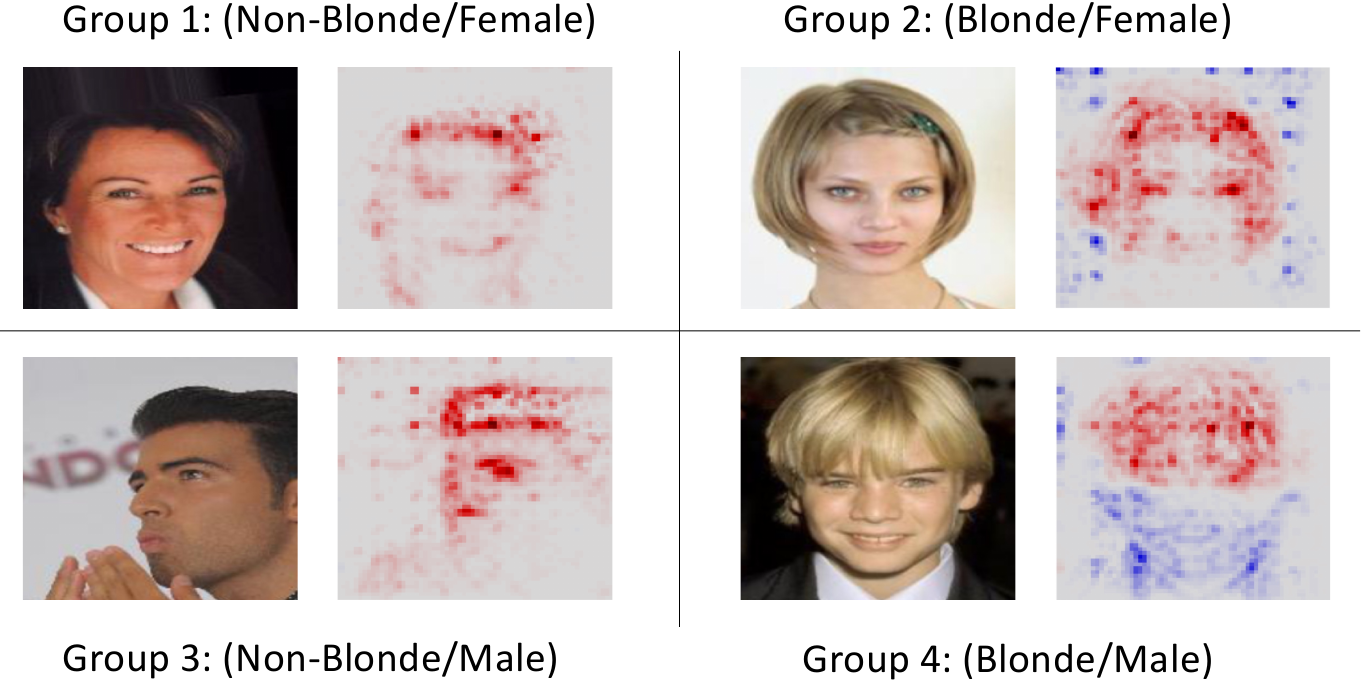}
    \vspace{-0.6cm}
    \caption{ExMap Heatmaps on CelebA: Each entry represents a group. The positive and negative attributions help interpret which features the model considers spurious (Blue), and which features are helpful (Red).}\vspace{-0.3cm}
    
    \label{fig:celeba}
\end{figure}
In this section, we present analysis and ablations along five axes: First, we demonstrate how the clustering of heatmaps is more useful than the clustering of features. Second, we demonstrate the usefulness of the ExMap representations when compared to ERM with respect to the classification task. Third, we provide more insight into what the learned clusters by ExMap capture in the data. Fourth, we demonstrate that ExMap is robust to the choice of clustering method, by performing an ablation on the clustering method using k-means instead of spectral clustering. Finally, to demonstrate that ExMap is strategy-agnostic, we use JTT as a retraining strategy using ExMap pseudo-labels, and are able to demonstrate robust performance with respect to JTT trained on true validation labels. 

\subsection{The benefit of heatmaps over features}
In this section we add more insight into why leveraging heatmaps for worst group robustness is more useful over features, as for example done in GEORGE. Specifically, we illustrate how heatmap based clustering mitigates reliance on the image background, the spurious attribute in the Waterbirds dataset. The results illustrate a common intuition - Explainability heatmaps highlight only the features relevant for prediction, ignoring those that are not. 
\paragraph{Circumventing background reliance}

Here, we present results on Waterbirds with the spurious attribute, i.e. background, removed. We call this variant Waterbirds (FG-Only), following \cite{Kirichenko2022LastLR}. Please refer to the supplementary section for examples. An effective group robustness method would not witness a sharp drop in test accuracy if the model does not rely on the background. In Table \ref{tab:fgonly}, we present these results. 

\begin{table}[t]
    \centering
    \resizebox{0.5\textwidth}{!}{
    \begin{tabular}{lccc}
        \toprule
        \textbf{Methods} & \textbf{Mean (FG-Only \%)} & \textbf{Mean (\%)} & \textbf{Drop}
        \color{Green}{$\downarrow$} \\
        \midrule
        ERM & 44.2 & 98.1 & 53.9 \\
        \rowcolor{gray!25}
        DFR & 64.7 & 94.6 & 29.9 \\
        GEORGE (DFR) & 73.2 & 96.5 & 23.3 \\
        DFR+ExMap (ours) & 78.5 & 96.0 & \textbf{17.5} \\
        \bottomrule
    \end{tabular}
    }
    \vspace{-3mm}\caption{Waterbirds (FG-Only). All methods exhibit a reliance on the background shortcut in Waterbirds, but ExMap posts the lowest drop, demonstrating its robustness.}\vspace{-3mm}
    \label{tab:fgonly}
\end{table}

We can clearly see that the heatmap clustering strategy mitigates the background reliance better than the feature based clustering strategy of GEORGE (lowest drop among all methods). This is also intuitive as the heatmap attributions focus on only the relevant features for prediction, discarding the rest (see Figure \ref{fig:heatmaps}).




\subsection{Qualitative Analysis}
\paragraph{ExMap improves explanations upon retraining}

We visualize the heatmaps and predictions of ERM and Exmap based DFR in Figure \ref{fig:heatmaps}. This is an image of the Waterbirds dataset that ERM misclassifies. This is reflected on the heatmap, as ERM fails to capture the relevant features. On the other hand, ExMap based DFR correctly classifies the image and focuses on the correct object region of interest (bird), instead of the spurious attribute (background).

\paragraph{ExMap improves Model Strategy}

\begin{table}[!t]
    \centering
    \resizebox{0.5\textwidth}{!}{
    \begin{tabular}{lcll}
        \toprule
        \textbf{Methods} & \textbf{Group Info} & \textbf{WGA(\%)} \color{Green}{$\uparrow$} & \textbf{Mean(\%)} \color{Green}{$\uparrow$} \\
        \midrule
        Base (ERM) & \xmark/\xmark & 76.8 & 98.1 \\
        \rowcolor{gray!25}
        DFR & \xmark/\cmark & 92.1 & 96.7 \\  
        \hline
        DFR+ExMap (SC) & \xmark/\xmark & 92.6 & 96.0 \\
        DFR+ExMap (KMeans) & \xmark/\xmark & 92.5 & 95.9 \\
        \bottomrule
    \end{tabular}
    }
    \vspace{-3mm}
    \caption{Worst group accuracy and mean accuracy on Waterbirds with two different clustering methods - kmeans and Spectral.}\vspace{-5mm}
    \label{tab:ablate1}
\end{table}

In Figure \ref{fig:celeba}, each entry represents a particular group. The positive and negative relevance scores correspond to the features that the model considers relevant and spurious respectively. ExMap uncovers the strategy used to make the prediction: In all four groups, we see ExMap helps the model uncover the hair color as a strategy. In fact, in Group 4, the model also learns that the facial features (Gender) are \textit{negatively} associated with the prediction task (Hair Color), which is what we desire from our method. The model has learned the shortcut between man and not-blonde hair, hence ExMap uncovers the negative relevance in the face, effectively uncovering this shortcut. These examples impart a notion of interpretability to our results, as we are able to explain why the model made a particular prediction, and what shortcuts are uncovered.


\subsection{Ablation Analysis}
\paragraph{Robustness to choice of clustering method}
Our proposed method does not depend on any particular clustering algorithm. Although we used spectral clustering, one can also use the simpler K-means \cite{kmeans} to capture the clusters for pseudo-labelling. In Table \ref{tab:ablate1}, we present the results on Waterbirds. We are able to demonstrate that there is no significant difference in the worst group robustness performance for the clustering method we choose\footnote{Note, empirical results illustrated that k-means results were robust to the number of clusters, $K$, given that $K$ was chosen sufficiently large.}. Both improve upon the base ERM and DFR models, and hence, both are useful. Thus, ExMap is more about demonstrating the usefulness of a heatmap clustering pseudo-labelling module rather than the specifics of the clustering method itself. 

\paragraph{Robustness to choice of learning strategy}
All the results presented until now focus on the DFR backbone for shortcut mitigation. We mention previously that ExMap is strategy-agnostic, meaning that it can be applied to any off-the shelf method in use today. In Table \ref{tab:ablate2}, we show the results after applying ExMap to the JTT method on Waterbirds. We demonstrate similar performance to using JTT (originally uses validation labels) simply by using the pseudo labels proposed by ExMap. Additionally, we are able to improve over ERM's poor worst group accuracy as well. 

        
        

\begin{table}[!t]
    \centering
    \resizebox{0.5\textwidth}{!}{
    \begin{tabular}{lcll}
        \toprule
        \textbf{Methods} & \textbf{Group Info} & \textbf{WGA(\%)} \color{Green}{$\uparrow$} & \textbf{Mean(\%)} \color{Green}{$\uparrow$} \\
        \midrule
        Base (ERM) & \xmark/\xmark & 41.1 & 95.9 \\
        \hline
        \rowcolor{gray!25}
        DFR & \xmark/\cmark &  92.1 & 96.7\\
        
        DFR+ExMap (ours) & \xmark/\xmark & 92.6 & 96.0
        \\
        \hline
        \rowcolor{gray!25}
        JTT & \xmark/\cmark & 86.7 & 93.3 \\  
        
        JTT+ExMap (ours) & \xmark/\xmark & 86.9 & 90.0\\
        \bottomrule
    \end{tabular}
    }
    \vspace{-3mm}\caption{Worst group and mean accuracy on Waterbirds for two different retraining strategies - JTT and DFR.}\vspace{-5mm}
    \label{tab:ablate2}
\end{table}

\section{Conclusion}
\label{sec:conclusion}

The group robustness paradigm for deep learning classifiers raises important questions for when deep learning models succeed, but more importantly, \textit{when they fail}. However, most of current research focuses on the setting where group labels are available. This assumption is impractical for real-world scenarios, where the underlying spurious correlations in the data may not be known apriori. While recent work investigating unsupervised group robustness mechanisms have shown promise, we show that further improvements are possible. In our work, we propose ExMap, where we cluster explainable heatmaps to generate pseudo-labels for the validation data. These pseudo-labels are then used on off-the-shelf group robustness learning mechanisms in use today. In addition to showing why using heatmaps over raw features is useful in this setting, our results demonstrate the efficacy of this approach on a range of benchmark datasets, in both the single and multi-shortcut settings. We are able to further close the gap to supervised counterparts, and outperform partially supervised and unsupervised baselines. Finally, ExMap opens up interesting avenues to further leverage explainability heatmaps in group robust learning.

\section*{Acknowledgements}
        \vspace{-.2cm}
        This work was financially supported by the Research Council of Norway (RCN) FRIPRO grant no.\ 315029. It was further funded through the RCN Centre for Research-based Innovation funding scheme (Visual Intelligence, grant no.\ 309439), and Consortium Partners.


{
    \small
    \bibliographystyle{ieeenat_fullname}
    \bibliography{main}

\begin{thebibliography}{40}
\providecommand{\natexlab}[1]{#1}
\providecommand{\url}[1]{\texttt{#1}}
\expandafter\ifx\csname urlstyle\endcsname\relax
  \providecommand{\doi}[1]{doi: #1}\else
  \providecommand{\doi}{doi: \begingroup \urlstyle{rm}\Url}\fi

\bibitem[Arjovsky et~al.(2019)Arjovsky, Bottou, Gulrajani, and Lopez-Paz]{Arjovsky2019InvariantRM}
Mart{\'i}n Arjovsky, L{\'e}on Bottou, Ishaan Gulrajani, and David Lopez-Paz.
\newblock Invariant risk minimization.
\newblock \emph{ArXiv}, abs/1907.02893, 2019.

\bibitem[Asgari et~al.(2022)Asgari, Khani, Khani, Gholami, Tran, Mahdavi-Amiri, and Hamarneh]{Taghanaki2022MaskTuneMS}
Saeid Asgari, Aliasghar Khani, Fereshte Khani, Ali Gholami, Linh Tran, Ali Mahdavi-Amiri, and Ghassan Hamarneh.
\newblock Masktune: Mitigating spurious correlations by forcing to explore.
\newblock \emph{Advances in Neural Information Processing Systems}, 2022.

\bibitem[Bach et~al.(2015)Bach, Binder, Montavon, Klauschen, M{\"u}ller, and Samek]{Bach2015OnPE}
Sebastian Bach, Alexander Binder, Gr{\'e}goire Montavon, Frederick Klauschen, Klaus-Robert M{\"u}ller, and Wojciech Samek.
\newblock On pixel-wise explanations for non-linear classifier decisions by layer-wise relevance propagation.
\newblock \emph{PLOS ONE}, 10, 2015.

\bibitem[Bahng et~al.(2019)Bahng, Chun, Yun, Choo, and Oh]{Bahng2019LearningDR}
Hyojin Bahng, Sanghyuk Chun, Sangdoo Yun, Jaegul Choo, and Seong~Joon Oh.
\newblock Learning de-biased representations with biased representations.
\newblock \emph{International Conference on Machine Learning}, 2019.

\bibitem[Brendel and Bethge(2018)]{brendel2018approximating}
Wieland Brendel and Matthias Bethge.
\newblock Approximating cnns with bag-of-local-features models works surprisingly well on imagenet.
\newblock \emph{International Conference on Learning Representations}, 2018.

\bibitem[Chen et~al.(2019)Chen, Carvalho, Baracaldo, Ludwig, Edwards, Lee, Molloy, and Srivastava]{Chen2018DetectingBA}
Bryant Chen, Wilka Carvalho, Nathalie Baracaldo, Heiko Ludwig, Benjamin Edwards, Taesung Lee, Ian Molloy, and Biplav Srivastava.
\newblock Detecting backdoor attacks on deep neural networks by activation clustering.
\newblock \emph{Workshop on Artificial Intelligence Safety}, 2019.

\bibitem[Duchi et~al.(2023)Duchi, Hashimoto, and Namkoong]{duchi2023distributionally}
John Duchi, Tatsunori Hashimoto, and Hongseok Namkoong.
\newblock Distributionally robust losses for latent covariate mixtures.
\newblock \emph{Operations Research}, 71\penalty0 (2):\penalty0 649--664, 2023.

\bibitem[Duchi et~al.(2021)Duchi, Glynn, and Namkoong]{duchi2018statistics}
John~C Duchi, Peter~W Glynn, and Hongseok Namkoong.
\newblock Statistics of robust optimization: A generalized empirical likelihood approach.
\newblock \emph{Mathematics of Operations Research}, 46\penalty0 (3):\penalty0 946--969, 2021.

\bibitem[Geirhos et~al.(2018)Geirhos, Rubisch, Michaelis, Bethge, Wichmann, and Brendel]{Geirhos2018ImageNettrainedCA}
Robert Geirhos, Patricia Rubisch, Claudio Michaelis, Matthias Bethge, Felix~A Wichmann, and Wieland Brendel.
\newblock Imagenet-trained cnns are biased towards texture; increasing shape bias improves accuracy and robustness.
\newblock \emph{International Conference on Learning Representations}, 2018.

\bibitem[Geirhos et~al.(2020)Geirhos, Jacobsen, Michaelis, Zemel, Brendel, Bethge, and Wichmann]{Geirhos2020ShortcutLI}
Robert Geirhos, J{\"o}rn-Henrik Jacobsen, Claudio Michaelis, Richard~S. Zemel, Wieland Brendel, Matthias Bethge, and Felix Wichmann.
\newblock Shortcut learning in deep neural networks.
\newblock \emph{Nature Machine Intelligence}, 2:\penalty0 665 -- 673, 2020.

\bibitem[Guo et~al.(2017)Guo, Li, and Ling]{Guo2017LIMELI}
Xiaojie Guo, Yu Li, and Haibin Ling.
\newblock Lime: Low-light image enhancement via illumination map estimation.
\newblock \emph{IEEE Transactions on Image Processing}, 26:\penalty0 982--993, 2017.

\bibitem[Izmailov et~al.(2022)Izmailov, Kirichenko, Gruver, and Wilson]{Izmailov2022OnFL}
Pavel Izmailov, Polina Kirichenko, Nate Gruver, and Andrew~G Wilson.
\newblock On feature learning in the presence of spurious correlations.
\newblock \emph{Advances in Neural Information Processing Systems}, 35:\penalty0 38516--38532, 2022.

\bibitem[Kirichenko et~al.(2022)Kirichenko, Izmailov, and Wilson]{Kirichenko2022LastLR}
Polina Kirichenko, Pavel Izmailov, and Andrew~Gordon Wilson.
\newblock Last layer re-training is sufficient for robustness to spurious correlations.
\newblock \emph{International Conference on Learning Representations}, 2022.

\bibitem[Kohlbrenner et~al.(2020)Kohlbrenner, Bauer, Nakajima, Binder, Samek, and Lapuschkin]{Kohlbrenner2019TowardsBP}
Maximilian Kohlbrenner, Alexander Bauer, Shinichi Nakajima, Alexander Binder, Wojciech Samek, and Sebastian Lapuschkin.
\newblock Towards best practice in explaining neural network decisions with lrp.
\newblock \emph{International Joint Conference on Neural Networks}, pages 1--7, 2020.

\bibitem[Lapuschkin et~al.(2019)Lapuschkin, W{\"a}ldchen, Binder, Montavon, Samek, and M{\"u}ller]{Lapuschkin2019UnmaskingCH}
Sebastian Lapuschkin, Stephan W{\"a}ldchen, Alexander Binder, Gr{\'e}goire Montavon, Wojciech Samek, and Klaus-Robert M{\"u}ller.
\newblock Unmasking clever hans predictors and assessing what machines really learn.
\newblock \emph{Nature Communications}, 10, 2019.

\bibitem[Lee et~al.(2023)Lee, Yao, and Finn]{Lee2023DiversifyAD}
Yoonho Lee, Huaxiu Yao, and Chelsea Finn.
\newblock Diversify and disambiguate: Out-of-distribution robustness via disagreement.
\newblock \emph{International Conference on Learning Representations}, 2023.

\bibitem[Levy et~al.(2020)Levy, Carmon, Duchi, and Sidford]{Lvy2020LargeScaleMF}
Daniel Levy, Yair Carmon, John~C Duchi, and Aaron Sidford.
\newblock Large-scale methods for distributionally robust optimization.
\newblock \emph{Advances in Neural Information Processing Systems}, 33:\penalty0 8847--8860, 2020.

\bibitem[Li et~al.(2023)Li, Evtimov, Gordo, Hazirbas, Hassner, Ferrer, Xu, and Ibrahim]{Li2022AWD}
Zhiheng Li, I. Evtimov, Albert Gordo, Caner Hazirbas, Tal Hassner, Cristian~Cant{\'o}n Ferrer, Chenliang Xu, and Mark Ibrahim.
\newblock A whac-a-mole dilemma: Shortcuts come in multiples where mitigating one amplifies others.
\newblock \emph{IEEE/CVF Conference on Computer Vision and Pattern Recognition}, pages 20071--20082, 2023.

\bibitem[Liu et~al.(2021)Liu, Haghgoo, Chen, Raghunathan, Koh, Sagawa, Liang, and Finn]{Liu2021JustTT}
Evan~Z Liu, Behzad Haghgoo, Annie~S Chen, Aditi Raghunathan, Pang~Wei Koh, Shiori Sagawa, Percy Liang, and Chelsea Finn.
\newblock Just train twice: Improving group robustness without training group information.
\newblock \emph{International Conference on Machine Learning}, pages 6781--6792, 2021.

\bibitem[Liu et~al.(2015)Liu, Luo, Wang, and Tang]{liu2015faceattributes}
Ziwei Liu, Ping Luo, Xiaogang Wang, and Xiaoou Tang.
\newblock Deep learning face attributes in the wild.
\newblock \emph{International Conference on Computer Vision}, 2015.

\bibitem[Lokhande et~al.(2022)Lokhande, Sohn, Yoon, Udell, Lee, and Pfister]{Lokhande2022TowardsGR}
Vishnu~Suresh Lokhande, Kihyuk Sohn, Jinsung Yoon, Madeleine Udell, Chen-Yu Lee, and Tomas Pfister.
\newblock Towards group robustness in the presence of partial group labels.
\newblock \emph{ICML 2022: Workshop on Spurious Correlations, Invariance and Stability}, 2022.

\bibitem[Lundberg and Lee(2017)]{Lundberg2017AUA}
Scott~M Lundberg and Su-In Lee.
\newblock A unified approach to interpreting model predictions.
\newblock \emph{Advances in neural information processing systems}, 30, 2017.

\bibitem[Montavon et~al.(2019)Montavon, Binder, Lapuschkin, Samek, and M{\"u}ller]{Montavon2019LayerWiseRP}
Gr{\'e}goire Montavon, Alexander Binder, Sebastian Lapuschkin, Wojciech Samek, and Klaus-Robert M{\"u}ller.
\newblock Layer-wise relevance propagation: an overview.
\newblock \emph{Explainable AI: interpreting, explaining and visualizing deep learning}, pages 193--209, 2019.

\bibitem[Na et~al.(2010)Na, Xumin, and Yong]{kmeans}
Shi Na, Liu Xumin, and Guan Yong.
\newblock Research on k-means clustering algorithm: An improved k-means clustering algorithm.
\newblock \emph{Third International Symposium on Intelligent Information Technology and Security Informatics}, pages 63--67, 2010.

\bibitem[Nam et~al.(2021)Nam, Kim, Lee, and Shin]{Nam2022SpreadSA}
Junhyun Nam, Jaehyung Kim, Jaeho Lee, and Jinwoo Shin.
\newblock Spread spurious attribute: Improving worst-group accuracy with spurious attribute estimation.
\newblock \emph{International Conference on Learning Representations}, 2021.

\bibitem[Nam et~al.(2020)Nam, Cha, Ahn, Lee, and Shin]{Nam2020LearningFF}
Jun~Hyun Nam, Hyuntak Cha, Sungsoo Ahn, Jaeho Lee, and Jinwoo Shin.
\newblock Learning from failure: De-biasing classifier from biased classifier.
\newblock \emph{Neural Information Processing Systems}, 2020.

\bibitem[Oren et~al.(2019)Oren, Sagawa, Hashimoto, and Liang]{Oren2019DistributionallyRL}
Yonatan Oren, Shiori Sagawa, Tatsunori~B Hashimoto, and Percy Liang.
\newblock Distributionally robust language modeling.
\newblock \emph{Empirical Methods in Natural Language Processing and the 9th International Joint Conference on Natural Language Processing}, pages 4227--4237, 2019.

\bibitem[Sagawa et~al.(2019)Sagawa, Koh, Hashimoto, and Liang]{Sagawa2019DistributionallyRN}
Shiori Sagawa, Pang~Wei Koh, Tatsunori~B Hashimoto, and Percy Liang.
\newblock Distributionally robust neural networks.
\newblock \emph{International Conference on Learning Representations}, 2019.

\bibitem[Schulth et~al.(2022)Schulth, Berghoff, and Neu]{Schulth2022DetectingBP}
Lukas Schulth, Christian Berghoff, and Matthias Neu.
\newblock Detecting backdoor poisoning attacks on deep neural networks by heatmap clustering.
\newblock \emph{ArXiv}, abs/2204.12848, 2022.

\bibitem[Selvaraju et~al.(2016)Selvaraju, Das, Vedantam, Cogswell, Parikh, and Batra]{Selvaraju2016GradCAMVE}
Ramprasaath~R. Selvaraju, Abhishek Das, Ramakrishna Vedantam, Michael Cogswell, Devi Parikh, and Dhruv Batra.
\newblock Grad-cam: Visual explanations from deep networks via gradient-based localization.
\newblock \emph{International Journal of Computer Vision}, 128:\penalty0 336 -- 359, 2016.

\bibitem[Singla and Feizi(2021)]{DBLP:journals/corr/abs-2110-04301}
Sahil Singla and Soheil Feizi.
\newblock Causal imagenet: How to discover spurious features in deep learning?
\newblock \emph{CoRR}, abs/2110.04301, 2021.

\bibitem[Sohoni et~al.(2020)Sohoni, Dunnmon, Angus, Gu, and R{\'e}]{Sohoni2020NoSL}
Nimit Sohoni, Jared Dunnmon, Geoffrey Angus, Albert Gu, and Christopher R{\'e}.
\newblock No subclass left behind: Fine-grained robustness in coarse-grained classification problems.
\newblock \emph{Advances in Neural Information Processing Systems}, 33:\penalty0 19339--19352, 2020.

\bibitem[Sohoni et~al.(2022)Sohoni, Sanjabi, Ballas, Grover, Nie, Firooz, and Re]{Sohoni2021BARACKPS}
Nimit~Sharad Sohoni, Maziar Sanjabi, Nicolas Ballas, Aditya Grover, Shaoliang Nie, Hamed Firooz, and Christopher Re.
\newblock Barack: Partially supervised group robustness with guarantees.
\newblock \emph{ICML 2022: Workshop on Spurious Correlations, Invariance and Stability}, 2022.

\bibitem[Sundararajan et~al.(2017)Sundararajan, Taly, and Yan]{Sundararajan2017AxiomaticAF}
Mukund Sundararajan, Ankur Taly, and Qiqi Yan.
\newblock Axiomatic attribution for deep networks.
\newblock \emph{International Conference on Machine Learning}, 2017.

\bibitem[Tsirigotis et~al.(2024)Tsirigotis, Monteiro, Rodriguez, Vazquez, and Courville]{Tsirigotis2023GroupRC}
Christos Tsirigotis, Joao Monteiro, Pau Rodriguez, David Vazquez, and Aaron~C Courville.
\newblock Group robust classification without any group information.
\newblock \emph{Advances in Neural Information Processing Systems}, 36, 2024.

\bibitem[Wah et~al.(2011)Wah, Branson, Welinder, Perona, and Belongie]{WahCUB_200_2011}
C. Wah, S. Branson, P. Welinder, P. Perona, and S. Belongie.
\newblock Caltech-ucsd-birds-2011.
\newblock Technical Report CNS-TR-2011-001, California Institute of Technology, 2011.

\bibitem[Wu et~al.(2023)Wu, Yuksekgonul, Zhang, and Zou]{Wu2023DiscoverAC}
Shirley Wu, Mert Yuksekgonul, Linjun Zhang, and James Zou.
\newblock Discover and cure: Concept-aware mitigation of spurious correlation.
\newblock \emph{International Conference on Machine Learning}, 2023.

\bibitem[Zhang et~al.(2022)Zhang, Sohoni, Zhang, Finn, and Re]{Zhang2022}
Michael Zhang, Nimit~S Sohoni, Hongyang~R Zhang, Chelsea Finn, and Christopher Re.
\newblock Correct-n-contrast: a contrastive approach for improving robustness to spurious correlations.
\newblock \emph{International Conference on Machine Learning}, 162:\penalty0 26484--26516, 2022.

\bibitem[Zhou et~al.(2017)Zhou, Lapedriza, Torralba, and Oliva]{Zhou2016PlacesAI}
Bolei Zhou, Agata Lapedriza, Antonio Torralba, and Aude Oliva.
\newblock Places: An image database for deep scene understanding.
\newblock \emph{Journal of Vision}, 17\penalty0 (10):\penalty0 296--296, 2017.

\bibitem[Zhou et~al.(2021)Zhou, Ma, Michel, and Neubig]{Zhou2021ExaminingAC}
Chunting Zhou, Xuezhe Ma, Paul Michel, and Graham Neubig.
\newblock Examining and combating spurious features under distribution shift.
\newblock \emph{International Conference on Machine Learning}, pages 12857--12867, 2021.

\end{thebibliography}


\begin{thebibliography}{14}
\providecommand{\natexlab}[1]{#1}
\providecommand{\url}[1]{\texttt{#1}}
\expandafter\ifx\csname urlstyle\endcsname\relax
  \providecommand{\doi}[1]{doi: #1}\else
  \providecommand{\doi}{doi: \begingroup \urlstyle{rm}\Url}\fi

\bibitem[Ahmadian et~al.(2019)Ahmadian, Epasto, Kumar, and Mahdian]{ahmadian2019clustering}
Sara Ahmadian, Alessandro Epasto, Ravi Kumar, and Mohammad Mahdian.
\newblock Clustering without over-representation.
\newblock \emph{International Conference on Knowledge Discovery \& Data Mining}, pages 267--275, 2019.

\bibitem[Arjovsky et~al.(2019)Arjovsky, Bottou, Gulrajani, and Lopez-Paz]{Arjovsky2019InvariantRM}
Mart{\'i}n Arjovsky, L{\'e}on Bottou, Ishaan Gulrajani, and David Lopez-Paz.
\newblock Invariant risk minimization.
\newblock \emph{ArXiv}, abs/1907.02893, 2019.

\bibitem[Bach et~al.(2015)Bach, Binder, Montavon, Klauschen, M{\"u}ller, and Samek]{Bach2015OnPE}
Sebastian Bach, Alexander Binder, Gr{\'e}goire Montavon, Frederick Klauschen, Klaus-Robert M{\"u}ller, and Wojciech Samek.
\newblock On pixel-wise explanations for non-linear classifier decisions by layer-wise relevance propagation.
\newblock \emph{PLOS ONE}, 10, 2015.

\bibitem[Bercea et~al.(2019)Bercea, Gro{\ss}, Khuller, Kumar, R{\"o}sner, Schmidt, and Schmidt]{bercea2019cost}
Ioana~O Bercea, Martin Gro{\ss}, Samir Khuller, Aounon Kumar, Clemens R{\"o}sner, Daniel~R Schmidt, and Melanie Schmidt.
\newblock On the cost of essentially fair clusterings.
\newblock \emph{Approximation, Randomization, and Combinatorial Optimization. Algorithms and Techniques (APPROX/RANDOM 2019)}, 2019.

\bibitem[Chhabra et~al.(2021)Chhabra, Masalkovait{\.e}, and Mohapatra]{chhabra2021overview}
Anshuman Chhabra, Karina Masalkovait{\.e}, and Prasant Mohapatra.
\newblock An overview of fairness in clustering.
\newblock \emph{IEEE Access}, 9:\penalty0 130698--130720, 2021.

\bibitem[Kirichenko et~al.(2022)Kirichenko, Izmailov, and Wilson]{Kirichenko2022LastLR}
Polina Kirichenko, Pavel Izmailov, and Andrew~Gordon Wilson.
\newblock Last layer re-training is sufficient for robustness to spurious correlations.
\newblock \emph{International Conference on Learning Representations}, 2022.

\bibitem[Li et~al.(2023)Li, Evtimov, Gordo, Hazirbas, Hassner, Ferrer, Xu, and Ibrahim]{Li2022AWD}
Zhiheng Li, I. Evtimov, Albert Gordo, Caner Hazirbas, Tal Hassner, Cristian~Cant{\'o}n Ferrer, Chenliang Xu, and Mark Ibrahim.
\newblock A whac-a-mole dilemma: Shortcuts come in multiples where mitigating one amplifies others.
\newblock \emph{IEEE/CVF Conference on Computer Vision and Pattern Recognition}, pages 20071--20082, 2023.

\bibitem[Liu et~al.(2021)Liu, Haghgoo, Chen, Raghunathan, Koh, Sagawa, Liang, and Finn]{Liu2021JustTT}
Evan~Z Liu, Behzad Haghgoo, Annie~S Chen, Aditi Raghunathan, Pang~Wei Koh, Shiori Sagawa, Percy Liang, and Chelsea Finn.
\newblock Just train twice: Improving group robustness without training group information.
\newblock \emph{International Conference on Machine Learning}, pages 6781--6792, 2021.

\bibitem[Liu et~al.(2015)Liu, Luo, Wang, and Tang]{liu2015faceattributes}
Ziwei Liu, Ping Luo, Xiaogang Wang, and Xiaoou Tang.
\newblock Deep learning face attributes in the wild.
\newblock \emph{International Conference on Computer Vision}, 2015.

\bibitem[Sagawa et~al.(2019)Sagawa, Koh, Hashimoto, and Liang]{Sagawa2019DistributionallyRN}
Shiori Sagawa, Pang~Wei Koh, Tatsunori~B Hashimoto, and Percy Liang.
\newblock Distributionally robust neural networks.
\newblock \emph{International Conference on Learning Representations}, 2019.

\bibitem[Seo et~al.(2022)Seo, Lee, and Han]{seo2022unsupervised}
Seonguk Seo, Joon-Young Lee, and Bohyung Han.
\newblock Unsupervised learning of debiased representations with pseudo-attributes.
\newblock \emph{IEEE/CVF Conference on Computer Vision and Pattern Recognition}, pages 16742--16751, 2022.

\bibitem[Sohoni et~al.(2020)Sohoni, Dunnmon, Angus, Gu, and R{\'e}]{Sohoni2020NoSL}
Nimit Sohoni, Jared Dunnmon, Geoffrey Angus, Albert Gu, and Christopher R{\'e}.
\newblock No subclass left behind: Fine-grained robustness in coarse-grained classification problems.
\newblock \emph{Advances in Neural Information Processing Systems}, 33:\penalty0 19339--19352, 2020.

\bibitem[Wah et~al.(2011)Wah, Branson, Welinder, Perona, and Belongie]{WahCUB_200_2011}
C. Wah, S. Branson, P. Welinder, P. Perona, and S. Belongie.
\newblock Caltech-ucsd-birds-2011.
\newblock Technical Report CNS-TR-2011-001, California Institute of Technology, 2011.

\bibitem[Zhou et~al.(2017)Zhou, Lapedriza, Torralba, and Oliva]{Zhou2016PlacesAI}
Bolei Zhou, Agata Lapedriza, Antonio Torralba, and Aude Oliva.
\newblock Places: An image database for deep scene understanding.
\newblock \emph{Journal of Vision}, 17\penalty0 (10):\penalty0 296--296, 2017.

\end{thebibliography}
}
\end{document}



\clearpage
\setcounter{page}{1}
\maketitlesupplementary


In this supplementary material, we present additional details about the following:

\begin{itemize}
    \item  The datasets used - C-MNIST, Waterbirds, CelebA, Urbancars, Urbancars single shortcut variants, Waterbirds (FG-Only).
    \item Experimental Setup - The details on the heatmap extraction and clustering phase in ExMap. 
    \item Additional results providing further intuition on how ExMap captures underlying group information. 
    \item More results on the robustness of our method with standard errors.
    \item The connection between group robustness and fair clustering.
    \item Limitations and Societal Impact.
\end{itemize}



\section{Datasets}
\label{sec:rationale}
%
%

We present the number of examples from each group for all the datasets, and the process of generating them. For C-MNIST, we used the same setup as in \cite{Arjovsky2019InvariantRM}. For Waterbirds and CelebA, we use the same setup as in \cite{Kirichenko2022LastLR, Liu2021JustTT}. For Urbancars we use the same setup as in \cite{Li2022AWD}. 

\subsection{C-MNIST}

We create a dataset where we have control of the number of elements in each group and what the spurious attribute is.

The Colored-MNIST dataset is a synthetic dataset based on the well-known MNIST.
The MNIST dataset is a collection of several thousands of examples of handwritten digits (0-9). The images are single-channelled (black and white) and have a size of 28x28 pixels, and are accompanied by a label giving the ground truth. 

We use the original data split, 60000 train and 10000 test. Since the original dataset does not have a validation set, we use the last 10000 images of the training set as the validation set.

We convert the dataset into a 2 class problem by modifying the task. This is done by simply going over to classify the numbers as smaller or equal to 4 ($y=0 \,:\,\, \text{value}<=4$), and larger than 4 ($y=1 \,:\,\, \text{value}>4$). To create the spurious attributes we make use of colors. Red is used as the first spurious attribute ($s=0 \,:\,\, \text{RGB}=(255, 0, 0)$), and green is used as the second spurious attribute ($s=1 \,:\,\, \text{RGB}=(0,255, 0)$). Naturally, the images will need to be made 3-channeled to account for this change.

As we are interested in combating spurious correlations we create the dataset in a way such that there are correlations between the classes and spurious attributes. We use $99\%$ correlation. That means that $99\%$ of images from one class will have the same colour, while the remaining $1\%$ will have the other colour. The amount of correlation was deliberately chosen so that ERM worst group accuracy is low. \autoref{tab:CMNIST} shows the number of images in each group for each split.

\begin{table}[t]
\centering
\begin{tabular}{|c|c|p{1cm}|p{1cm}|p{1cm}|p{1cm}|}
\hline
Split & Total Data & \multicolumn{4}{c|}{Groups} \\
\cline{3-6}
 & & Group 0 (y=0, s=0) & Group 1 (y=0, s=1) & Group 2 (y=1, s=0) & Group 3 (y=1, s=1) \\
\hline
Train & 50,000 & 254 & 25,284 & 24,231 & 231 \\
Val & 10,000 & 45 & 5,013 & 4,893 & 49 \\
Test & 10,000 & 48 & 5,091 & 4,815 & 46 \\
\hline 
\end{tabular}
\caption{Data splits in the Colored-MNIST dataset.}
\label{tab:CMNIST}
\end{table}

\subsection{Waterbirds}

Waterbirds \cite{Sagawa2019DistributionallyRN} is a synthetic dataset created with the purpose of testing a model's reliance on background. The dataset consists of RGB images depicting different types of birds on different types of backgrounds. The different types of birds are divided into 2 classes, landbirds ($y=0$) and waterbirds ($y=1$). The different backgrounds are also divided into 2 and represent the spurious attributes of this dataset: land background ($s=0$) and water background ($s=1$). The group distributions across the different splits are presented in \autoref{tab:WATERBIRDS}.

The Waterbirds dataset is created by using 2 other datasets, the Caltech-UCSD Birds-200-2011 (CUB) dataset \cite{WahCUB_200_2011} and the Places dataset \cite{Zhou2016PlacesAI}.
The CUB dataset contains images of birds labelled by species and their segmentation masks. To construct the Waterbirds dataset the labels in the CUB dataset are split into 2 groups, where waterbirds are made up of seabirds (albatross, auklet, cormorant, frigatebird, fulmar, gull, jaeger, kittiwake, pelican, puffin, or tern) and waterfowls (gadwall, grebe, mallard, merganser, guillemot, or Pacific loon), while the remaining classes are labelled as landbirds.
The birds are cropped using the pixel-level segmentation masks and pasted onto a water background (categories: ocean or natural lake) or land background (categories: bamboo forest or broadleaf forest) from the Places dataset.

The official train-test split of the CUB dataset is used, and $20\%$ of the training set is used to create the validation set.
The group distribution for the training set is such that most images ($95\%$) depict bird types with corresponding backgrounds, to represent a distribution that may arise from real-world data. This distribution turns the background into a spurious feature.
Take note that there is a distribution shift from the training split to the validation and test splits which are both more balanced, and include many more elements for the minority group. The creators of the dataset argue that they do this to more accurately gauge the performance of the minority groups, something that might be difficult if there are too few examples. They also do this to allow for easier hyperparameter tuning. 

\begin{table}[t]
\centering
\begin{tabular}{|c|c|p{1cm}|p{1cm}|p{1cm}|p{1cm}|}
\hline
Split & Total Data & \multicolumn{4}{c|}{Groups} \\
\cline{3-6}
 & & Group 0 (y=0, s=0) & Group 1 (y=0, s=1) & Group 2 (y=1, s=0) & Group 3 (y=1, s=1) \\
\hline
Train & 4,795 & 3,498 & 184 & 56 & 1,057 \\
Val & 1,199 & 467 & 466 & 133 & 133 \\
Test & 5,794 & 2,255 & 2,255 & 642 & 642 \\
\hline 
\end{tabular}
\caption{Data splits in the Waterbirds dataset.}
\label{tab:WATERBIRDS}
\end{table}

\subsection{Celeb-A}
CelebA here is a reference to a part of the CelebA celebrity face dataset \cite{liu2015faceattributes} that was introduced by \cite{Sagawa2019DistributionallyRN} as a group robustness dataset. From the original dataset, the feature \textit{Blond\_Hair} is used as the class, meaning that the images are divided into people who are not blonde ($y=0$) and blonde ($y=1$). Meanwhile, as a spurious attribute, we use the feature \textit{Male} from the original dataset, which divides into female ($s=0$) and male ($s=1$).
The official train-val-test split of the CelebA dataset is used. Note in \autoref{tab:CELEBA}, that the splits are likely randomly created, which results in equally group-distributed splits. Across all splits the group (blonde, male) is the smallest.

This dataset tests for model reliance on strongly correlated features in a real-world dataset. Observe in \autoref{tab:CELEBA} that $g_3=(y=1, s=1)$ which represents blonde males is severely underrepresented compared to the other groups, hence we expect the model to learn gender as a spurious feature for the class blonde.

\begin{table}[ht]
\centering
\begin{tabular}{|c|c|p{1cm}|p{1cm}|p{1cm}|p{1cm}|}
\hline
Split & Total Data & Group 0 (y=0, s=0) & Group 1 (y=0, s=1) & Group 2 (y=1, s=0) & Group 3 (y=1, s=1) \\
\hline
Train & 162,770 & 71,629 & 66,874 & 22,880 & 1,387 \\
Val & 19,867 & 8,535 & 8,276 & 2,874 & 182 \\
Test & 19,962 & 9,767 & 7,535 & 2,480 & 180 \\
\hline
\end{tabular}
\caption{Data splits in the CelebA dataset.}
\label{tab:CELEBA}
\end{table}

\subsection{Urbancars}

We use Urbancars, as proposed by \cite{Li2022AWD}. There are 4000 images per target class, i.e. 8000 images in total. The target class is the car type (country/urban), while the two shortcuts are the background type (country/urban), and co-occurring object (country/urban). For the exact list of the cars, objects, and background, please see \cite{Li2022AWD}. 

\subsection{Urbancars single shortcut variants}



The original Urbancars data has eight group combinations due to two classes, and two shortcuts (Background and Co-Occurring object). For the single shortcut variants, we merge the 4 extra groups for one particular shortcut, to leave 4 groups for the other. For example: To create Urbancars (BG), we merge the 4 groups from the other shortcut (CoObj), to create four groups containing the single shortcut of background for each of the two classes. A similar procedure is adopted to create Urbancars (CoObj). 

\subsection{Waterbirds (FG-Only)}

\begin{figure}[ht]
    \centering
    \includegraphics[width=0.35\textwidth]{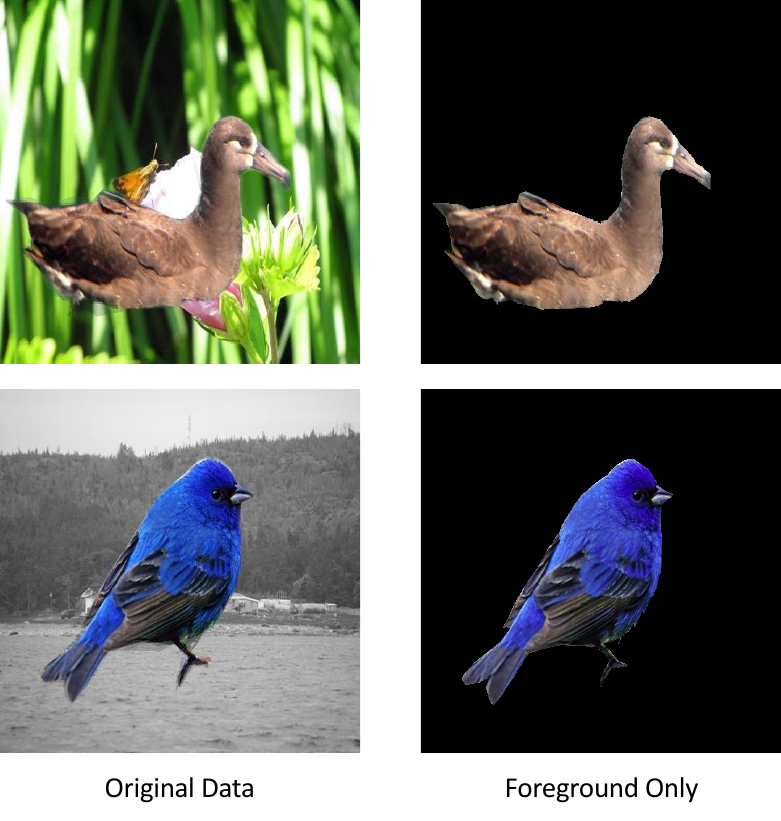}
    \caption{Waterbirds (FG Only)}
    \label{fig:fgonly}
\end{figure}

This dataset is created to evaluate how well the trained models circumvent background reliance on the Waterbirds dataset, since background is the shortcut in the data. We remove the backgrounds in all the images only on the test set. In Figure \ref{fig:fgonly}, we present some examples. 

\section{Experimental Setup}

In this section, we present more details on the heatmap extraction phase, clustering choice, and the hyperparameters used.

\begin{table*}[!t]
    \centering
    \resizebox{0.9\linewidth}{!}{%
    \begin{tabular}{lcllllll}
        \toprule
        \textbf{Methods} & \textbf{Group Info} & \multicolumn{2}{c}{\textbf{C-MNIST}} & \multicolumn{2}{c}{\textbf{Urbancars (BG)}} & \multicolumn{2}{c}{\textbf{Urbancars (CoObj)}} \\
        \cmidrule(lr){3-4} \cmidrule(lr){5-6} \cmidrule(lr){7-8}
        & Train/Val & WGA(\%)\color{Green}{$\uparrow$} & Mean(\%) & WGA(\%)\color{Green}{$\uparrow$} & Mean(\%) & WGA(\%)\color{Green}{$\uparrow$} & Mean(\%) \\
        \midrule
        Base (ERM) & \xmark/\xmark & 39.6 & 99.3 & 55.6 & 90.2 & 50.8 & 92.7\\
        \hline
        GEORGE (DFR) & \xmark/\xmark & 71.7\(\pm0.1\) & 95.2\(\pm0.3\) & 69.1\(\pm0.9\) & 83.6\(\pm1.0\) & 76.9\(\pm0.9\) & 91.4\(\pm1.0\) \\
        DFR+ExMap (ours) & \xmark/\xmark & \textbf{72.5\(\pm0.2\)} & 94.9\(\pm0.3\) & \textbf{71.4\(\pm0.8\)} & 93.2\(\pm0.2\) & \textbf{79.2\(\pm0.7\)} & 93.2\(\pm0.3\) \\
        \bottomrule
    \end{tabular}}
    \caption{Group/mean test accuracy with std. Results over 5 runs.}
    \label{tab:suppresult-george}
\end{table*}

\subsection{Heatmap Extraction}

Following SPRAY \cite{Bach2015OnPE}, which reports good results across different downsizing of heatmaps, we sweep predominantly over the following downsizings: [224, 112, 100, 56, 28, 14, 7, 5, 3]. The downsizing of heatmaps additionally helps in speeding up the clustering process and mitigating potential out-of-memory issues.
\begin{figure}[!t]
    \centering
    \includegraphics[width=0.5\textwidth]{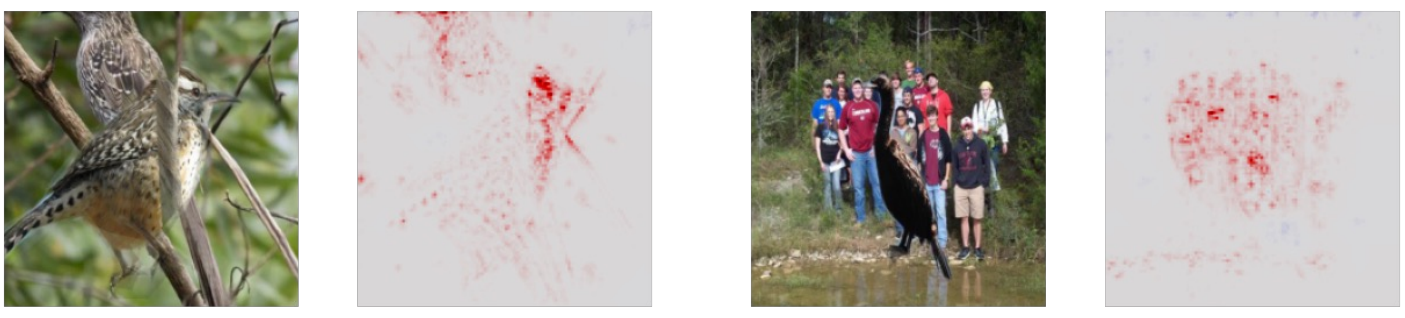}
    \caption{ExMap based misclassifications on challenging examples (Waterbirds): In each of these images, the object of interest (bird) is co-habited by dominant peripheral objects such as humans and other birds. These situations are challenging for the classifier to discern the relevant object from the irrelevant ones.}
    \label{fig:failure}
\end{figure}
\subsection{Clustering choice}

Since ExMap if flexible to the choice of clustering algorithm, we experiment with spectral clustering, UMap reduced KMeans \cite{Kirichenko2022LastLR}, and KMeans. We use the eigengap heuristic with spectral to automatically choose the number of clusters, and sweep over different cluster sizes for the KMeans based methods.
We look for the largest gap in among the first 10 eigenvalues. Otherwise we test for 2-15 clusters for kmeans (overclustering as practiced in \cite{Sohoni2020NoSL}).

\subsection{Hyperparameters}
We use the same hyperparameters for DFR and JTT as in the original papers \cite{Kirichenko2022LastLR, Liu2021JustTT}. 

For DFR, we perform the following steps:
\begin{itemize}
    \item Given (pseudo)group labels we create a retraining set by subsampling each group to the size of the smallest group. These are then used to retrain the last layer.  After being passed through the feature extractor, each sample is normalised based on the data used to retrain the last layer.
    \item Similar to \cite{Kirichenko2022LastLR}, we use logistic regression with L1-loss.
    \item The strength of L1 is swept over [1.0, 0.7, 0.3, 0.1, 0.07, 0.03, 0.01]. The sweep is performed by randomly splitting the retraining dataset in 2, and performing retraining with one half and evaluating the performance with the other. This is performed 5 times with different splits and the best strength is chosen based on highest worst (pseudo)group accuracy.
    \item When L1 strength has been selected, we retrain using the whole retrain set. This is performed 20 times with different subsamplings. The weights from each subsampling are averaged (this is viable according to DFR authors) to yield the final last layer weights. The normalisation of data is also averaged across the 20 runs.
\end{itemize}

For the ERM model, we perform the following steps:
\begin{itemize}
    \item We use Resnet-18 for CMNIST, Resnet-50 for the others. We start with imagenet-pretrained Resnet-50 similar to previous work as it was observed to perform better. For all settings we replace the final fully connected layer to reflect the nature of our problems, i.e. 2 classes.
    \item Learning rate: 3e-3, weight decay: 1e-4, cosine learning rate scheduler. 
    \item Batch size:We use batch size of 32 for Waterbirds and Urbancars, 100 for CelebA, and 128 for C-Mnist.
    \item Epochs: We train for 100 epochs on Waterbirds and Urbancars, 20 for  CelebA, and 10 for C-Mnist.
    \item We use early stopping using the best mean (weighted) validation accuracy
\end{itemize}

For GEORGE, we perform the following steps:
\begin{itemize}
    \item Acquire feature extractor (base ERM) outputs. 
    \item Max normalise features. 
    \item Cluster features as exmap or using UMAP+kmeans. We use 2 dimensions for the UMAP reduction, and high number of clusters (overclustering regime following \cite{Sohoni2020NoSL}).
\end{itemize}

\begin{table}[!t]
    \scriptsize
    \centering
    \setlength{\tabcolsep}{2pt}
    \resizebox{0.4\textwidth}{!}{
        \begin{tabular}{lcc}
            \toprule
            \textbf{Method} & \multicolumn{2}{c}{\textbf{Accuracy (\%)}} \\
            & \textbf{Waterbirds} & \textbf{CelebA} \\
            & WGA / Mean & WGA / Mean \\
            \midrule
            Base (ERM) & 76.8 / 98.1 & 41.1 / 95.9 \\
          
            GEORGE (DFR) & 91.7 $\pm$ 0.2 / 96.5 $\pm$ 0.1 & 83.3 $\pm$ 0.2 / 89.2 $\pm$ 0.2 \\
            DFR+ExMap & \textbf{92.5 $\pm$ 0.1} / 96.0 $\pm$ 0.3 & \textbf{84.4 $\pm$ 0.5} / 91.8 $\pm$ 0.2 \\
            \bottomrule\vspace{-0.6cm}
        \end{tabular}
    }
    \caption{Group / mean test accuracy with std. Results over 5 runs.\vspace{-0.6cm}}
    \label{tab:suppresultrebut}
\end{table}

\begin{table*}[!t]
    \centering
    \setlength{\tabcolsep}{12pt}
    \resizebox{0.9\linewidth}{!}{\begin{tabular}{lcllll}
    
        \toprule
        \textbf{Methods} & \textbf{Group Info} & \multicolumn{2}{c}{\textbf{Waterbirds}} & \multicolumn{2}{c}{\textbf{CelebA}} \\
        \cmidrule(lr){3-4} \cmidrule(lr){5-6}
        & Train/Val & WGA(\%)\color{Green}{$\uparrow$} & Mean(\%) & WGA(\%)\color{Green}{$\uparrow$} & Mean(\%) \\
        \midrule
        Base (ERM) & \xmark/\xmark & 76.8 & 98.1 & 41.1 & 95.9\\
        \hline

        BPA & \xmark/\xmark & 71.3 & 87.1 & 83.3 & 90.1 \\
        DFR+ExMap (ours) & \xmark/\xmark & \textbf{92.5} & 96.0 & \textbf{84.4} & 91.8 \\
        \bottomrule
    \end{tabular}}
    \caption{Comparison with Fair Clustering: Worst group and mean accuracy on Waterbirds and CelebA. }
    \label{tab:suppresult}
\end{table*}



\section{Capturing of Group Information} 
\begin{figure}[tp]
    \centering\vspace{-0.15cm}
    \includegraphics[width=0.4\textwidth]{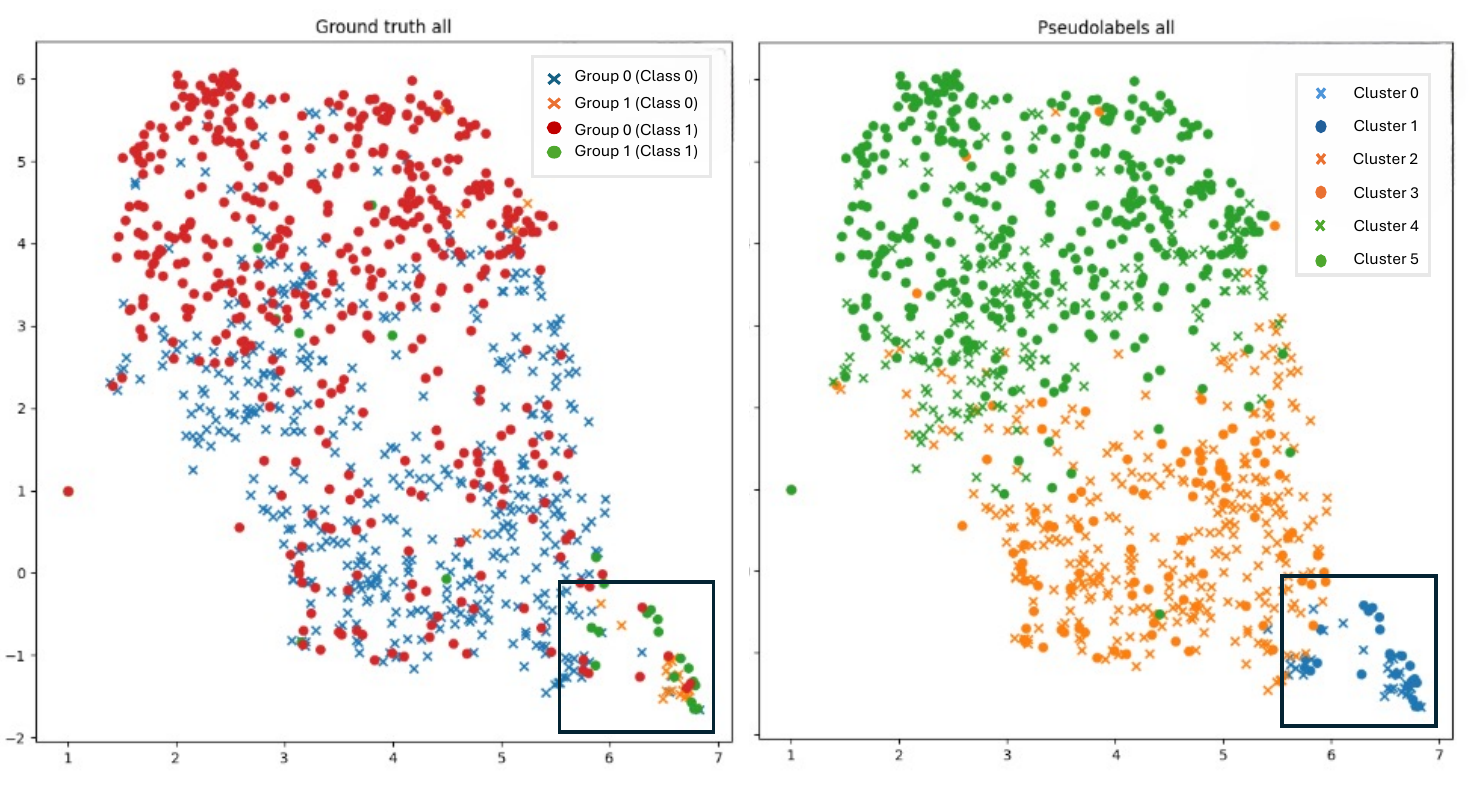}
    \vspace{-0.4cm}\caption{Groups in UrbanCars: (Left) Ground truth group labels per class. We observe minority groups (the spurious correlations) in the highlighted bottom right corner. (Right) Pseudo-labels learned by ExMap based clustering reveals a similar overall structure, conserving the dominant groups (green and yellow), while capturing the minority groups (blue cross and circle) as well.}\vspace{-0.4cm}
\label{fig:cluster}
\end{figure}

In addition to why ExMap representations are better for downstream group robustness over raw classifier features, we are also interested in what kind of group information the ExMap representations capture. The advantage of heatmaps are that they capture only the relevant features, while previous approaches that cluster in the feature space are prone to be effected by features that are irrelevant for the final prediction. To further substantiate our findings, we generate additional results to demonstrate that ExMap indeed captures the underlying group information. In Figure \ref{fig:cluster}, we plot the pseudo-labels for UrbanCars (CoObj) after ExMap based clustering. ExMap captures both the dominant groups and the minority groups in the dataset, as indicated by the pseudo-labels learned. We also note that ExMap does not necessarily learn the same number of groups as in the ground truth data, since this information is assumed unavailable. The key observation from this figure is that ExMap is successful in identifying the dominant and minority group structure in the data. The group robust learner (such as DFR) can then sample across these groups in a balanced manner while retraining, leading to mitigation against spurious correlations. 

\section{Robustness Analysis}

Our results in Table 1 and Table 2 in the main text are presented as the average of five runs. To illustrate the robustness of the compared approaches, we further provide the standard deviation for the ExMap and the main competitor in Table \ref{tab:suppresult-george} and Table \ref{tab:suppresultrebut}. We observe that results are robust across runs.


\section{Connections to Fair Clustering}
Given the close relationship between group robustness and the domain of fair clustering~\cite{bercea2019cost, ahmadian2019clustering, chhabra2021overview}, we briefly comment on their connection and the potential of the insights of ExMap in the fair clustering setting. The domains of fair clustering and group robustness differ slightly, with the former aiming to improve mean accuracy independent of sensitive attributes, while the latter aim to maximize worst group accuracy. Therefore, there is a natural connection between these two research areas. Sensitive attributes in fair clustering can be regarded as a special type of spurious correlation, causally unrelated to the task. Recent work in fair clustering has therefore adopted some of the insights from the field of group robustness \cite{seo2022unsupervised}. However, these approaches adopt a GEORGE inspired approach (cluster in raw features space), which we demonstrate to be sub-optimal in the context of group robustness. While an in-depth exploration of this is out-of-scope for this work, it could present an interesting avenue of future work. In Table \ref{tab:suppresult}, we present the ExMap results on Waterbirds and CelebA with respect to the method introduced in \cite{seo2022unsupervised}. 

\section{Limitations and Societal Impact}
There are certain intuitive failure cases where the ExMap approach is not as efficient. This occurs when the images themselves are quite challenging to discern the objects of interest (the class), from other peripheral objects in the scene. In Figure \ref{fig:failure}, we present some examples of misclassifications by ExMap based DFR. In these images, we can see that the object of interest (bird), is co-habited by other dominant objects in the scene, such as humans and other birds. This creates an exceptionally challenging task for the classifier to discern the relevant features for the task. We recognise the need for robustness across challenging examples in datasets as motivation for future work. With regard to social impact, we recognise that model robustness to spurious correlations is an important first step in ensuring fair, transparent, and reliable AI that can be deployed in safety critical domains in the real world. Elucidating why models classify as they do, and specific failure cases uncovers shortcomings in exclusively choosing mean test accuracy as a metric. As a result, probing models for their weaknesses is as important as exemplifying their strengths. 

{
    \small
    \bibliographystyle{ieeenat_fullname}
    \bibliography{main}
}